\documentclass[journal ]{new-aiaa}
\usepackage[utf8]{inputenc}
\usepackage{textcomp}

\usepackage{xspace}
\makeatletter
\DeclareRobustCommand\onedot{\futurelet\@let@token\@onedot}

\def\@onedot{\ifx\@let@token.\else.\null\fi\xspace}

\def\eg{\emph{e.g}\onedot} 
\def\ie{\emph{i.e}\onedot}

\def\etal{\emph{et al}\onedot}
\makeatother

\usepackage{comment}
\usepackage{xcolor,hyperref}
\usepackage{tikz}
\usetikzlibrary{calc}
\usetikzlibrary{intersections}
\usetikzlibrary{3d,calc,positioning,backgrounds,arrows.meta,decorations.markings,intersections,fit}
\usetikzlibrary{matrix}

\usepackage{cleveref}
\usepackage{algorithm}
\usepackage{algpseudocode}
\usepackage{amsmath}
\usepackage{pifont} 
\newcommand{\vmark}{\ding{51}} 
\newcommand{\xmark}{\ding{55}} 

\usepackage{nicematrix}
\usepackage{graphicx}
\usepackage{amsmath}
\usepackage[version=4]{mhchem}
\usepackage{siunitx}
\usepackage{multirow}
\usepackage{makecell}

\usepackage{array}

\usepackage{graphicx}
\usepackage{array}
\usepackage[export]{adjustbox}

\usepackage{adjustbox}

\newcommand{\imagecellnew}[2]{%
  \adjustbox{max size={#1}{#1}, valign=c, trim=0 0 0 0, clip}{%
    \includegraphics[width=#1,height=#1]{#2}%
  }%
}

\usepackage{longtable,tabularx}
\title{NeRF-based Spacecraft Reconstruction from Monocular Imagery Under Illumination Variability and Pose Uncertainty}

\author{Antoine Legrand \footnote{PhD Researcher, UCLouvain, ICTEAM, Dept. Electrical Engineering; KU Leuven, Dept. Electrical Engineering; KU Leuven, Flanders Make@KU Leuven; and Aerospace Engineer, Aerospacelab, (antoine.legrand@uclouvain.be).}}
\affil{ICTEAM, UCLouvain, Louvain-la-Neuve, 1348, Belgium}
\affil{ESAT, KU Leuven, Leuven, 3001, Belgium}
\affil{Flanders Make@KU Leuven, Leuven, 3001, Belgium}
\affil{Aerospacelab, Mont-Saint-Guibert, 1435, Belgium}
\author{Renaud Detry \footnote{Associate Professor, KU Leuven, Dept. Electrical Engineering; KU Leuven, Dept. Mechanical Engineering. KU Leuven; Flanders Make@KU Leuven}}
\affil{ESAT, KU Leuven, Leuven, 3001, Belgium}
\affil{MECH, KU Leuven, Leuven, 3001, Belgium}
\affil{Flanders Make@KU Leuven, Leuven, 3001, Belgium}
\author{Christophe De Vleeschouwer \footnote{Research Director, UCLouvain, ICTEAM, Dept. Electrical Engineering.}}
\affil{ICTEAM, UCLouvain, Louvain-la-Neuve, 1348, Belgium}

\begin{document}

\maketitle

\begin{abstract}
    Autonomous rendezvous and proximity operations around uncooperative, unknown spacecraft are critical for active debris removal and on-orbit servicing missions. A key component of such operations is the offline reconstruction of a 3D model of the target from a set of 2D images. This task is challenging due to two main factors. First, in-orbit illumination conditions exhibit considerable variability, and change rapidly over time. Second, the inaccuracy of pose information in the images, results in 3D reconstruction uncertainty. To overcome these challenges, we propose to extend Neural Radiance Fields with per-image degrees of freedom: a learnable appearance embedding that captures the illumination conditions specific to each image, and an image-specific pose correction term that refines its noisy pose label to increase 3D consistency across images. These parameters add minimal complexity, as they are learned jointly with the NeRF, yet they substantially improve robustness to illumination variability and pose inaccuracies. We validate our approach on three image sets representative of in-orbit operations, demonstrating its effectiveness for offline reconstruction and highlighting its suitability for online reconstruction, an open problem in the field.
\end{abstract}

\section*{Nomenclature}

{\renewcommand\arraystretch{1.0}
\noindent\begin{longtable*}{@{}l @{\quad=\quad} l@{}}
$c$  & origin of a 3D ray (camera center) in the reference frame \\
$d$  & normalized unit direction of a 3D ray in the reference frame, $\|d\|_2 = 1$ \\
$I$  & image of dimensions $W \times H$ pixels \\
$M_\Phi$ & NeRF enabling image synthesis from a given camera pose, with learned weights $\Phi$ (see \Cref{sec_background}) \\
$N$ & number of frames \\
$q$  & relative orientation between camera and reference frame, represented as a right-handed quaternion \\
$r,g,b$ & RGB color components \\
$t$  & position of the camera center expressed in the reference frame \\
$\sigma$ & point density \\
\multicolumn{2}{@{}l}{Indices}\\
$a$ & image (frame) index \\
$i$ & pixel index along image width \\
$j$ & pixel index along image height \\
\multicolumn{2}{@{}l}{Modifiers}\\
$\hat{x}$ & raw estimate \\
$\tilde{x}$ &  refined estimate \\
$\delta x$	& correction term\\
\end{longtable*}}

\addtocounter{table}{-1}

\section{Introduction}
\label{sec_introduction}

Autonomous Rendezvous and Proximity Operations (RPOs)~\cite{goodman2006history,boyarko2011optimal} have gained in interest over recent years. They involve maneuvers of a chaser spacecraft in close proximity of a target spacecraft, possibly culminating with the docking of the chaser on the target. With the increase in autonomy enabled by the adoption of Artificial Intelligence, these operations could be conducted autonomously, without human intervention. Such autonomous operations form the critical part of a new generation of space missions such as On-Orbit Servicing (OOS)~\cite{long2007orbit,flores2014review}, which consists in inspecting~\cite{tweddle2015relative,nakka2022information}, refueling~\cite{dutta2012peer,medina2017towards}, repairing~\cite{hastings2006orbit,ellery2008case} or assembling~\cite{cheng2016orbit,saunders2017building} space assets, or Active Debris Removal (ADR)~\cite{liou2010controlling,bonnal2013active}, which aims at removing end-of-life satellites or space debris from their orbits~\cite{kessler2010kessler}.

A key capability required for OOS and ADR is the 3D reconstruction of an unknown target spacecraft from close-range, on-orbit, images depicting that target. While a wide range of sensors could be used to capture images of a target spacecraft such as LIDAR or stereo cameras, our work focuses on monocular cameras due to their lower cost, mass and power consumption, as well as their compactness which makes them well suited for less expensive missions. The reconstructed 3D model can be used to identify the best grasping parts of the target~\cite{mavrakis2021orbit,sah2024target}, to train or validate an AI-enabled Guidance Navigation and Control (GNC) pipeline~\cite{legrand2024leveraging}, or as a key building block for relative pose estimation~\cite{heintz2023spacecraft}. Finally, the proposed offline 3D reconstruction method could also serve as an internal representation for future online reconstruction approaches. Such methods, operating onboard a chaser spacecraft during proximity operations, would pave the way for fully autonomous operations on uncooperative and unknown space debris, ultimately supporting their removal.

3D reconstruction from 2D images has recently attracted a significant interest among the computer vision community. In particular, Novel View Synthesis (NVS), which aims at rendering novel views of a scene given a set of images depicting that scene, rapidly evolved through tools such as Neural Radiance Fields (NeRFs)~\cite{mildenhall2021nerf} or 3D-Gaussian Splatting (3D-GS)~\cite{kerbl20233d}.  The principle behind these NVS tools consists in learning a representation from a set of images depicting a scene under different viewpoints. This representation is then used to render views of that scene under unseen viewpoints.

While previous works explored the use of NVS-based 3D model reconstruction methods in the context of Rendezvous and Proximity Operations~\cite{mergy2021vision,heintz2023spacecraft,nguyen2024characterizing}, they validated their methods on over-controlled benchmarks which ignored two significant problems that arise when dealing with a real use case. First, they did not consider the challenging illumination conditions that occur in orbit. Apart from the Earth albedo, which dimly contributes to the spacecraft illumination, the target is directly illuminated by the Sun. Due to the lack of atmospheric diffusion, only the spacecraft parts that are lit by the Sun are visible. Furthermore, those parts may be over-exposed because of the specular reflections that are typical of the materials of which the spacecraft is made of. An additional challenge related to the illumination conditions are their variability. Indeed, due to the long times required to safely approach the target spacecraft, both chaser and target move significantly on their orbits. As a result, the illumination conditions may dramatically vary during the operation. Second, Novel View Synthesis-based reconstruction methods rely on a set of images along their relative poses. Previous works assume a perfect knowledge of these pose labels. However, on a real use-case, they are unknown and can only be estimated, \eg, through human annotations. This uncertainty on the 6D pose labels results in an uncertainty on the 3D representation, and hence, in blurry 2D image reconstruction~\cite{klasson2024sources}.

Our contribution therefore consists in a novel view synthesis-based method for recovering a 3D representation of an unknown target spacecraft from a set of 2D images depicting that target. Unlike previous works, our method is designed to handle the illumination variability and the pose uncertainty that are inherent to real use cases. To handle the varying illumination conditions, our method adopts a Neural Radiance Field (NeRF)~\cite{mildenhall2021nerf} with in-the-wild abilities~\cite{martin2021nerf}, \ie, an appearance embedding is learned for each image in order to model implicitly the photometric variations between the training images. Similarly, for each image, we introduce a learnable pose correction term to correct the estimated pose label. By learning these terms along the NeRF, the labels are  progressively denoised, thereby reducing the uncertainty of the 3D representation, without affecting the model complexity. The proposed method is validated on a synthetic and two Hardware In the Loop sets that correspond to different orbital lighting scenarios.

The rest of this paper is structured as follows. \Cref{sec_related_works} positions our work with regards to previous arts and highlights their limitations. \Cref{sec_background} describes the rendering process of NeRFs as well as their training. \Cref{sec_method} presents our method for reconstructing a 3D model of an unknown target spacecraft from monocular images taken under different viewpoints with an emphasis on the handling of the illumination variability as well as the pose labels uncertainty. \Cref{sec_datasets} describes the datasets used in our experiments while \Cref{sec_experiments} validates our method as well as its components. Finally, \Cref{sec_conclusion} concludes.

\section{Related Works}
\label{sec_related_works}
    Offline 3D model reconstruction from monocular images aims to recover a 3D representation of an object given a set of images and their associated pose labels. Numerous studies have addressed this task in the context of orbital rendezvous and proximity operations. This section summarizes these works and highlights how our approach overcomes their limitations.
    
    Novel View Synthesis (NVS)-based methods have been widely adopted for this task. These methods learn a representation optimized to reconstruct the available images, which can then be used to render novel views of the target from unseen viewpoints. NVS approaches rely on either implicit or explicit representations. Implicit models describe the scene as a continuous function (\ie, a field), whereas explicit models represent it using geometric primitives. Consequently, implicit methods synthesize images via volumetric integration, while explicit methods use rasterization.
    
    Implicit representations, such as Neural Radiance Fields (NeRFs)~\cite{mildenhall2021nerf}, were the first considered for NVS-based 3D reconstruction. Mergy \etal~\cite{mergy2021vision} pioneered this direction by demonstrating NeRF’s ability to learn spacecraft geometry from monocular images. They also showed that training from unposed images is possible using Generative Radiance Fields (GRAFs)~\cite{schwarz2020graf}, although at a prohibitive computational cost due to an adversarial formulation. Subsequent works~\cite{caruso20233d,fu2024neural,forray2025joint,spacecraft_nerf_2025} leveraged advanced NeRF variants such as Instant-NGP~\cite{muller2022instant} and D-NeRF~\cite{pumarola2021d}, but their evaluations were limited to synthetic datasets or images captured under smooth illumination, neglecting the challenges of orbital lighting. Other studies addressed issues such as RGB-D input~\cite{han2024pose} or low-light and blurry in-orbit imagery~\cite{xu2025neural}, yet they still relied on synthetic data that poorly reflects real illumination conditions. In contrast, our method is specifically tailored for on-orbit lighting and is evaluated on images exhibiting these conditions.
    
    Explicit representations have recently gained attention for this task. Several works~\cite{nguyen2024characterizing,issitt2025optimal,park2025improved,de2025geometric} demonstrated that 3D Gaussian Splatting~\cite{kerbl20233d} can reconstruct spacecraft geometry from monocular synthetic images. Similarly, De Smijter \etal~\cite{de2025geometric} validated this capability for 3D Convex Splatting~\cite{held20253d}, which uses convex primitives. While explicit representations offer advantages such as lower computational cost and an explicit geometric description, we adopt NeRF-based models because their field-based formulation is better suited to handle adverse illumination conditions.
    
    Finally, several works aim to recover an abstract 3D structure of an unknown spacecraft from monocular images without relying on NVS representations~\cite{mathihalli2024dreamsat,park2024rapid,park2024improving,bates2025removing,huc2025fast}. Although conceptually appealing and promising for future developments, these approaches remain insufficiently robust to the diverse illumination conditions encountered in orbit. They are also less suited than NVS-based solution in dealing with rendering-based tasks. That is, NVS-based models enable rendering novel viewpoints, facilitating inspection and supporting the training of relative pose estimation networks~\cite{legrand2024leveraging,legrand2024domain}. They can also visualize decision cues~\cite{legrand2025nerf} or even perform pose estimation directly via inversion~\cite{heintz2023spacecraft}.

    Although pose inaccuracies have not been extensively studied in space applications, this issue has been addressed within the computer vision community. Unlike SLAM-based approaches~\cite{sucar2021imap,zhu2022nice,rosinol2023nerf,barad2024object}, which aim to reconstruct a scene from a stream of unposed images, or adversarial methods~\cite{schwarz2020graf,bian2023nope,fu2024cbarf} that reconstruct scenes from unposed image sets, our objective is different: we seek to learn the scene from images whose relative pose labels are slightly inaccurate. Inspired by prior works such as I-NeRF~\cite{yen2021inerf}, Parallel Inversion~\cite{lin2022parallel}, and BARF~\cite{lin2021barf}, our method leverages the differentiability of NeRFs to optimize pose correction terms applied to the inaccurate labels. This mechanism introduces no additional complexity while significantly enhancing reconstruction sharpness.

\section{Background: Novel View Synthesis Through NeRFs}
\label{sec_background}

    Novel View Synthesis (NVS) tools aim to generate images of a given scene from unseen viewpoints by learning a representation from a limited set of images captured under varying viewpoints. Various representations have been developed for NVS, including Neural Radiance Fields~\cite{mildenhall2021nerf} (NeRFs), 3D Gaussian Splatting~\cite{kerbl20233d}, or 3D Convex Splatting~\cite{held20253d}. In this paper, we adopt NeRFs due to their ray-tracing-based image generation process, which is well suited for handling challenging illumination conditions.

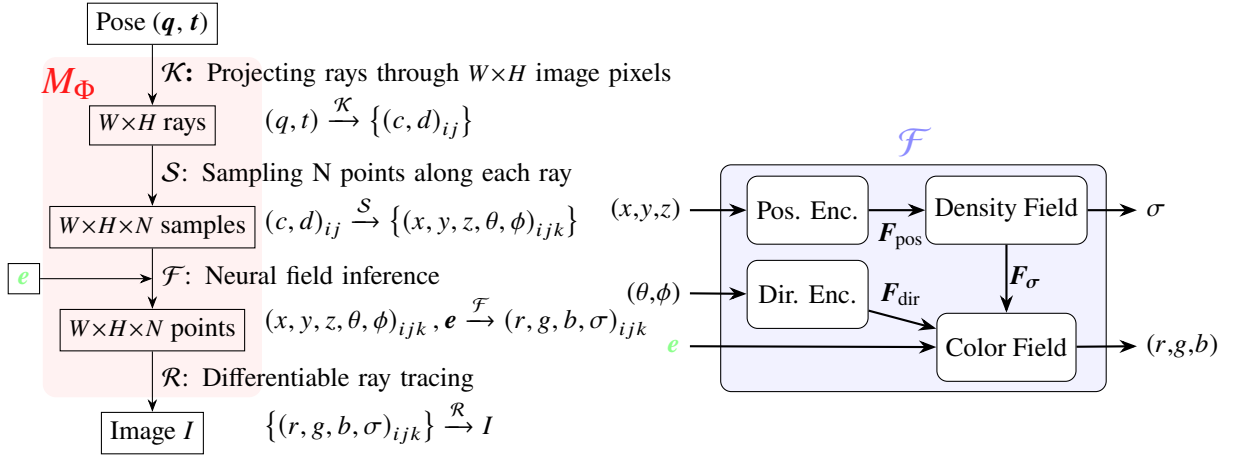
\begin{figure}[t]
    \begin{minipage}{.5\textwidth}
    \begin{flushleft}
    \begin{tikzpicture}[node distance=0.8cm, >=Stealth, align=left]
        \node (A) [rectangle, draw] {Pose ($\boldsymbol{q}$, $\boldsymbol{t}$)};
        \node (B) [rectangle, draw, below=of A] {{\small $W \!\times\! H$} rays};
        \node (C) [rectangle, draw, below=of B] {{\small$W \!\times\! H \!\times\! N$} samples};
        \node (D) [rectangle, draw, below=of C] {{\small$W \!\times \! H \! \times \! N$} points};
        \node (E) [rectangle, draw, below=of D] {Image $I$};
        
        
        \node (X3) [rectangle, draw] at ($(C)!0.5!(D) + (-1.7cm, 0)$) {\textcolor{green!45}{$\boldsymbol{e}$}};
        \draw[->] (X3.east) -- ($(C)!0.5!(D)$) node[midway, above, text width=3cm, align=center] {};

        \def\rectwidth{2.9}
        \def\rectheight{4.5}
        \begin{scope}[on background layer]
            \draw[rounded corners=5pt, fill=red!05, draw=none]
            ($(C) + (-\rectwidth/2, -\rectheight/2)$) 
            rectangle 
            ($(C) + (\rectwidth/2, \rectheight/2)$);
        \end{scope}
        
        \node (tempf) [left=0.8cm of B] {};
        \node (F) at ($(tempf)+(0.65cm,0.55cm)$) [text=red!90, font=\Large\bfseries] {$M_\Phi$};
        
        \def\distbox{1.35cm}
        
        \draw[->] (A) -- (B) node[very near start, right, text width=7.25cm, anchor=north west] {
            \textbf{$\mathcal{K}$:} Projecting rays through {\small$W \!\times\! H$} image pixels\\\vspace{0.075cm}
            $ \hspace{\distbox} \displaystyle (q,t) \xrightarrow{\mathcal{K}} \left\{ \left(c,d\right)_{ij}\right\}$
        };
        \draw[->] (B) -- (C) node[very near start, right, text width=7.25cm, anchor=north west]{
            $\mathcal{S}$: Sampling N points along each ray\\\vspace{0.075cm}
            $ \hspace{\distbox} \displaystyle \left(c,d\right)_{ij} \xrightarrow{\mathcal{S}} \left\{ \left(x,y,z,\theta,\phi\right)_{ijk}\right\}$
        };
        \draw[->] (C) -- (D) node[very near start, right, text width=7.25cm, anchor=north west]{
            $\mathcal{F}$: Neural field inference\\\vspace{0.075cm}
            $ \hspace{\distbox} \displaystyle \left(x,y,z,\theta,\phi\right)_{ijk}, \boldsymbol{e} \xrightarrow{\mathcal{F}} \left(r,g,b,\sigma\right)_{ijk}$
        };
        \draw[->] (D) -- (E) node[very near start, right, text width=7.25cm, anchor=north west] {
            $\mathcal{R}$: Differentiable ray tracing\\ \vspace{0.075cm}
            $ \hspace{\distbox} \displaystyle \left\{ \left(r,g,b,\sigma\right)_{ijk}\right\} \xrightarrow{\mathcal{R}} I $
        };
    \end{tikzpicture}
    \end{flushleft}
    \end{minipage}
    \hfill    
    \begin{minipage}{.5\textwidth}
    \begin{flushright}
    \vspace{0.7cm}
    \begin{tikzpicture}[node distance=0.7cm and 1.0cm, auto, >=Stealth]
        \tikzstyle{block} = [draw, fill=blue!05, rounded corners, minimum height=3.0cm, minimum width=5.1cm]
        \tikzstyle{component} = [draw, fill=white, rounded corners, minimum height=0.9cm, minimum width=1.6cm]
        \tikzstyle{input} = [coordinate]
        \tikzstyle{output} = [coordinate]
        \tikzstyle{arrow} = [->, thick]
        
        \node[block] (main) {};   
        \node[input, above=of main, yshift=-0.4cm, xshift=0.35cm] (F) {};  
        
        \node[input, left=of main, xshift=0.6cm, yshift=0.9cm] (A) {};
        \node[input, left=of main, xshift=0.6cm, yshift=-0.2cm] (B) {};
        \node[input, left=of main, xshift=0.6cm, yshift=-0.9cm] (E) {}; 
        
        \node[output, right=of main, xshift=-0.6cm, yshift=0.9cm] (C) {};
        \node[output, right=of main, xshift=-0.6cm, yshift=-0.9cm] (D) {};
        
        \node[component, right=0.75cm of A] (comp1) {Pos. Enc.};
        \node[component, below=0.2cm of comp1] (comp2) {Dir. Enc.};
        \node[component, right=0.75cm of comp1] (comp3) {Density Field};
        \node[component, below=0.9cm of comp3] (comp4) {Color Field};
        
        \node[left=0cm of F] {{\textcolor{blue!45}{\Large $\mathcal{F}$}}};
        \node[left=0cm of A] {($x$,$y$,$z$)};
        \node[left=0cm of B] {($\theta$,$\phi$)};
        \node[left=0cm of E] {\textcolor{green!45}{$\boldsymbol{e}$}};
        \node[right=0cm of C] {$\sigma$};
        \node[right=0cm of D] {($r$,$g$,$b$)};
        
        \node (tempa) [right=0.3cm of comp1] {};
        \node (a) at ($(tempa)+(0cm,-0.35cm)$) {$\boldsymbol{F_{\textnormal{pos}}}$};
        \node (tempb) [right=0.3cm of comp2] {};
        \node (b) at ($(tempb)+(0cm,0.0cm)$) {$\boldsymbol{F_{\textnormal{dir}}}$};
        \node (tempc) [below=0.3cm of comp3] {};
        \node (c) at ($(tempc)+(0.25cm,0cm)$) {$\boldsymbol{F_{\sigma}}$};
        
        \draw[arrow] (A) -- (comp1);
        \draw[arrow] (B) -- (comp2);
        \draw[arrow] (comp1) -- (comp3);
        \draw[arrow] (comp3) -- (C);
        \draw[arrow] (comp2) -- (comp4);
        \draw[arrow] (comp3) -- ++(0,-0.5) -| (comp4);
        \draw[arrow] (E) -- (comp4); 
        \draw[arrow] (comp4) -- (D);
    \end{tikzpicture}
    \end{flushright}
    \end{minipage}
\caption{\label{fig_back_nerf_rendering} Image generation through a Neural Radiance Field (see \cref{sec_background})}
\end{figure}

    \Cref{fig_back_nerf_rendering} illustrates the synthesis of an image \( I \) of dimensions \( W \times H \) from a camera pose \((q, t)\), where \( q \) denotes orientation and \( t \) denotes position, using a NeRF \( M_{\Phi} \). For each pixel, a ray is cast from the camera center through the pixel center. The ray is defined by its origin \( c_{ij} = t \) and unit direction \( \mathbf{d}_{ij} = R(q)\: \mathbf{d^{\textnormal{P}}}_{ij} \), where \( \mathbf{d^{\textnormal{P}}}_{ij} \) represents the direction of a ray originating from the center of a world-aligned camera and passing through pixel \( ij \). Along each ray, \( K \) points are sampled. For each sample, the 3D position \((x, y, z)\) and viewing direction \((\theta, \phi)\) are input to a neural field \( \mathcal{F} \), which predicts the density \( \sigma \) and color \((r, g, b)\) of the point. Finally, the pixel value is computed by aggregating the color and density of all sampled points along the ray using differentiable ray-tracing techniques.

    While many architectures exist, the neural field, which predicts the density and color of a point from its position and viewing angle, follows a common core principle. Each position ($x,y,z$) and viewing direction ($\theta$,$\phi$) are first encoded into position $F_{\textnormal{pos}}$ and direction features $F_{\textnormal{dir}}$, respectively.
    A Multilayer Perceptron (MLP), fed with the position features, then outputs the density $\sigma$ along with intermediate density features $F_\sigma$. These features are concatenated with the direction ones, and passed to a second MLP that predicts the color ($r$,$g$,$b$). 

    Key differences between NeRF implementations lie in the specific formulations of their neural field components. The original NeRF~\cite{mildenhall2021nerf} employed sinusoidal positional encoding for both spatial coordinates and viewing directions, combined with large MLPs to predict density and color. Although effective, this design led to slow training and rendering of novel views. Instant-NGP~\cite{muller2022instant} addresses these limitations by introducing a learnable multi-resolution hash grid for encoding positions, and spherical harmonics for encoding viewing directions. These innovations enable the use of smaller MLPs, significantly accelerating both training and inference while maintaining reconstruction quality.

    While such architectures are well suited for static scenes, they are not designed to handle illumination changes that naturally occur in unconstrained (in-the-wild) environments. To address this limitation, NeRF-in-the-wild~\cite{martin2021nerf} introduced the concept of appearance embeddings to model appearance discrepancies across training images. An appearance embedding is a learnable vector associated with a specific image, which is fed into the color MLP alongside the direction and density features. As the appearance embeddings for all training images are jointly learned with the NeRF, they capture the illumination specificities of their corresponding images.
    
    In this paper, we rely on a Neural Radiance Field based on the $K$-Planes~\cite{fridovich2023k} architecture, because it incorporates appearance embeddings while remaining computationally efficient through a positional encoding scheme similar to Instant-NeRF~\cite{muller2022instant}. The right part of \Cref{fig_back_nerf_rendering} illustrates the high-level architecture of the $K$-Planes-based neural field. Specifically, the positional and directional encoders consist of 3-plane interpolation and spherical harmonics, respectively.

    Since the entire image generation process is differentiable, NeRFs can be trained by backpropagating the photometric loss between the observed images and those rendered from the corresponding viewpoints, \ie,
\begin{equation}
    \Phi^{*} = \operatorname*{argmin}_{\Phi} \sum_{a=1}^{N} \mathcal{L}_\textnormal{Photo} \left(I_a, M_{\Phi}(q_a, t_a, e_a)\right)
\end{equation}
    where \( \Phi^{*} \) denotes the optimal NeRF weights. In practice, however, NeRFs are trained on batches composed of pixels randomly sampled from the training images along their corresponding rays, \ie,
\begin{equation}
    \Phi^{*} = \operatorname*{argmin}_{\Phi} \sum_{a=1}^{N} \sum_{i=1}^{W} \sum_{j=1}^{H} \mathcal{L}_\textnormal{Photo} \left(I_{aij}, \mathcal{R}(\mathcal{F}(\mathcal{S}(c_{aij}, d_{aij}, e_a)))\right)
\end{equation}
    This strategy ensures that each batch contains pixels from diverse viewpoints, promoting multi-view consistency and ultimately improving training convergence.

\section{Method}
\label{sec_method}
    This section describes our method for recovering an implicit representation of an uncooperative, unknown, target spacecraft from a set of monocular images depicting that target in orbit. \Cref{sec_problem_statement,sec_method_overview} describe the problem and present an overview of the method while \Cref{sec_method_pose_finetuning} details a key elements of our method, namely, the finetuning of the pose labels through learned correction terms.

\begin{figure}[t]
    \centering
    \includegraphics[width=0.95\textwidth]{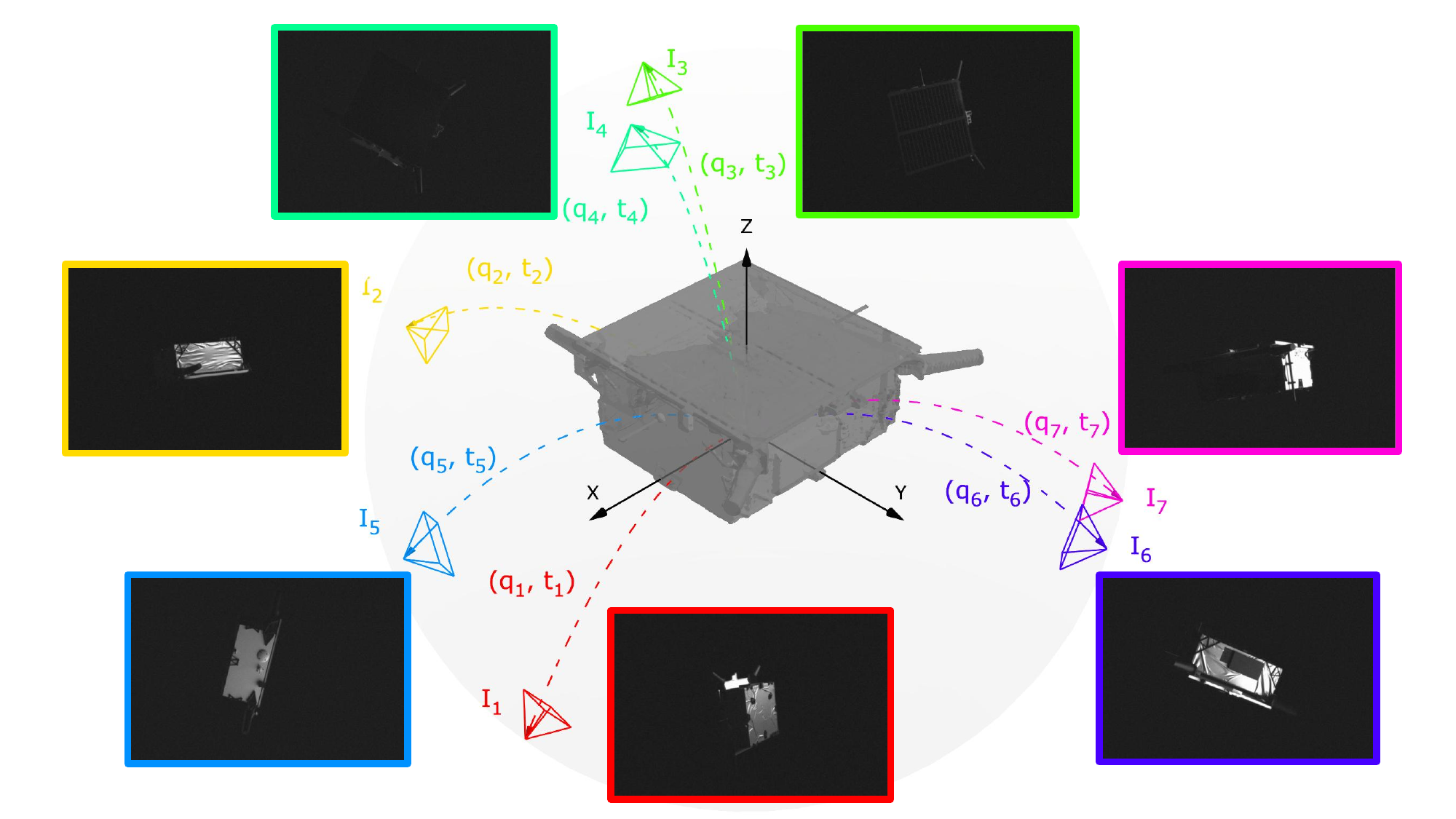}
    \caption{\label{fig_problem_overview}Problem Overview: our method learns a 3D, implicit, representation of a target spacecraft from a set of images taken under different viewpoints. The challenges are two-fold. First, due to the orbital lighting conditions, two images taken under close viewpoints may exhibit drastically different illuminations. Second, the relative pose between the camera and the target are unknown and can only be roughly estimated.}
\end{figure}

\subsection{Problem Statement}
\label{sec_problem_statement}

    \Cref{fig_problem_overview} illustrates the problem of recovering an implicit representation of an unknown target spacecraft $M_{\Phi}$ from a set of monocular images $\left\{ I_a\right\}_{a=1}^{N}$ taken under different viewpoints $\left\{ (q,t)_a\right\}_{a=1}^{N}$, and different orbital lighting conditions. Such a representation can then be used to generate novel views of the target taken under unseen viewpoints. 
    While this problem has partially been addressed~\cite{mergy2021vision}, \eg, through Neural Radiance Fields (see \Cref{sec_background}), two issues that arise on real use cases have not been addressed yet. 

    First, an underlying conventional assumption of implicit representations is the constant nature of the lighting conditions. All the training images are expected to be taken under similar illumination conditions. However, illumination conditions may abruptly and significantly change in orbit. For example, in \Cref{fig_problem_overview}, images $I_6$ and $I_7$ depict the same two perpendicular faces of the target. However, only one face is illuminated in each image while the other is completely under-exposed. These discrepancies between the training images seriously affect the training of the NeRF, and consequently, the quality of the model reconstructed by such an illumination-agnostic NeRF. 

    Second, NeRFs are usually trained on a set of images paired with their corresponding relative poses $(q,t)_a$, \ie, orientation $q$ and position $t$. However, in a real use-case, the ground-truth position and orientations are unknown, only the captured images are available. Relative poses can be estimated photogrammetrically, but the noise affecting these estimates remains an impediment. This uncertainty on the pose labels results in a inconsistent 3D representation. When such a representation is rendered, this leads to blur in the generated images.
    
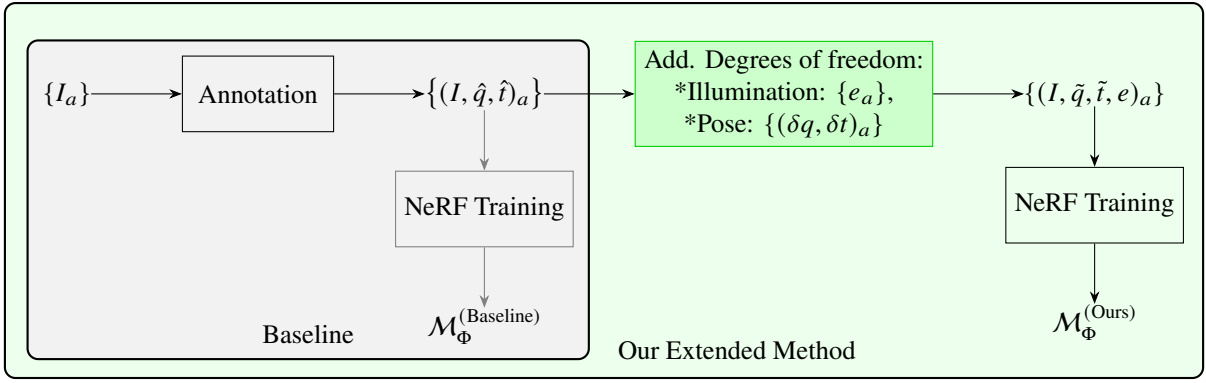
\begin{figure}[t]
\centering
\begin{tikzpicture}[node distance=1.5cm and 2cm, >=Stealth]

  \tikzstyle{block} = [rectangle, draw, minimum height=1cm, minimum width=2cm, align=center]
  \tikzstyle{textnode} = [inner sep=0pt, outer sep=0pt]
  \tikzstyle{grayblock} = [block, draw=gray, fill=gray!10]
  \tikzstyle{greenblock} = [block, draw=green!80!black, fill=green!20]  \tikzstyle{container} = [draw, thick, fill=green!07, inner sep=0.5cm, rounded corners]
  \tikzstyle{subcontainer} = [draw, thick, fill=gray!10, inner sep=0.2cm, rounded corners]
  
  \node[textnode] (n1) {$\left\{ I_a\right\}$};
  \node[block, right=1.2cm of n1] (b1) {Annotation};
  \node[textnode, right=1.2cm of b1] (n2) {$\left\{ (I,\hat{q},\hat{t})_a\right\}$};
  \node[greenblock, right=1.2cm of n2] (b2) {Add. Degrees of freedom: \\ *Illumination: $\left\{ e_a\right\}$, \\ *Pose: $\left\{ (\delta q,\delta t)_a\right\}$};
  \node[textnode, right=1.2cm of b2] (n3) {$\left\{ (I,\tilde{q},\tilde{t},e)_a\right\}$};

  \node[grayblock, below=0.8cm of n2] (b3) {NeRF Training};
  \node[block, below=0.8cm of n3] (b4) {NeRF Training};

  \node[textnode, below=0.8cm of b3] (n4) {$\mathcal{M}_{\Phi}^{(\textnormal{Baseline)}}$};
  \node[textnode, below=0.8cm of b4] (n5) {$\mathcal{M}_{\Phi}^{(\textnormal{Ours)}}$};

  \draw[->] (n1) -- (b1);
  \draw[->] (b1) -- (n2);
  \draw[->] (n2) -- (b2);
  \draw[->] (b2) -- (n3);
  \draw[->, gray] (n2) -- (b3);
  \draw[->, gray] (b3) -- (n4);
  \draw[->] (n3) -- (b4);
  \draw[->] (b4) -- (n5);
  
  \begin{pgfonlayer}{background}
    \node[container, fit=(n1)(n3)(n5)(b2), label={[anchor=south,yshift=-138pt, xshift=50pt] Our Extended Method}] (extended) {};
    
    \node[subcontainer, fit=(n1)(n2)(n4)(b1)(b3), label={[anchor=south,yshift=-118pt]Baseline}] (baseline) {};
    
  \end{pgfonlayer}
    
\end{tikzpicture}
\caption{\label{fig_overview_extension} Method Overview: our method extends the usual NeRF training paradigm by adding degrees of freedom to the model. While per-image appearance embeddings $\left\{e_a\right\}_{a=1}^{N}$ are learned to capture photometric variations between images taken under diverse illumination conditions, per-image pose correction terms $\left\{(\delta q, \mathbf{\delta t})_a\right\}_{a=1}^{N}$ are learned to refine the roughly annotated pose labels.}
\end{figure}

\subsection{Method Overview}
\label{sec_method_overview}

    As depicted in \Cref{fig_overview_extension}, our method recovers an implicit representation $M_\Phi$ from a set of images $\left\{ I_a\right\}_{a=1}^{N}$ by extending the NeRF-based paradigm. This extension addresses the two previously highlighted challenges, namely, the NeRF inability to model varying and harsh lighting conditions, as well as the 3D uncertainty inherent with the use of estimates of the relative poses of the training images. Both challenges are addressed through a conceptually simple, yet efficient, mechanism, which consists in increasing the degrees of freedom of the neural radiance field, with regard to both illumination variations as well as pose labels uncertainty, as illustrated in \Cref{fig_overview_extension}.
    
    To model the photometric variations between images taken under similar viewpoints but different illumination conditions, we rely on appearance embeddings~\cite{martin2021nerf}. Each embedding is associated to a given image and implicitly capture the illumination conditions that are specific to that image. This additional, per-image, degrees of freedom allows the model to handle challenging and varying illumination conditions.

    Similarly, as detailed in the next section, to mitigate the impact of the pose labels uncertainty, we learn for each image a pose correction term. This term, which encompasses the correction of both the camera orientation and position, is learned along the representation and is able to refine the coarse pose estimate. This reduces the uncertainty of the pose labels, which, in turn, improves the 3D representation consistency, and 2D rendering sharpness.

\subsection{Addressing Pose Uncertainty Trough Learnable Pose Correction Terms}
\label{sec_method_pose_finetuning}

The Neural Radiance Field $M_{\Phi}$ should normally be trained on ground-truth, \ie, perfectly accurate, pose labels, $\left\{(q_a, \boldsymbol{t}_a)\right\}_{a=1}^N$. 
However, due to errors inherent to human annotations, the NeRF is trained on pseudo-pose labels $\left\{(\hat{q}_a,\hat{\boldsymbol{t}}_a)\right\}_{a=1}^N$.  
The epistemic uncertainty on the pose labels results in an uncertainty on the 3D reconstruction~\cite{klasson2024sources}, which is ultimately responsible for a blur observed in the images generated through this model. 

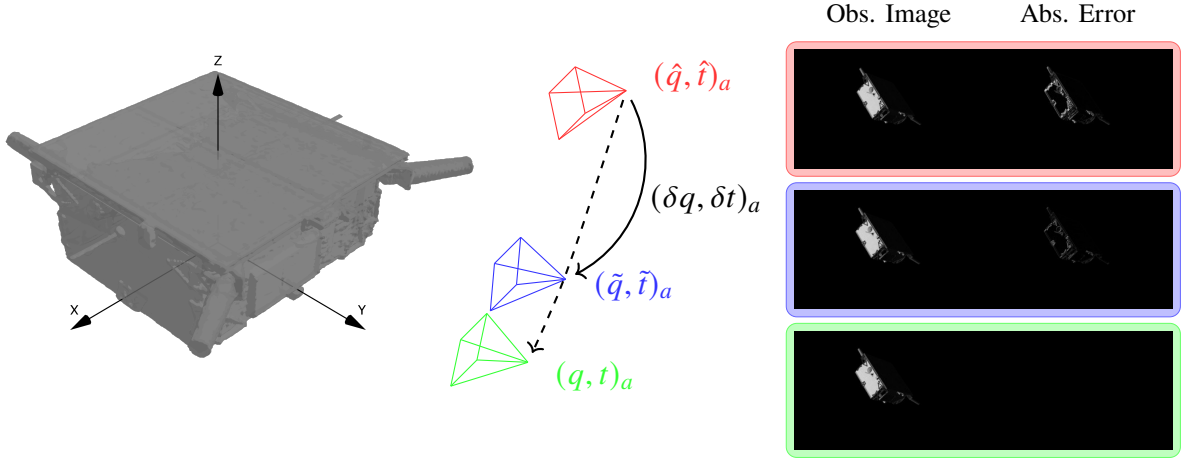
\begin{figure}[t]
\centering
\begin{tikzpicture}[scale=1]

\pgfmathsetmacro{\camX}{0.9}
\pgfmathsetmacro{\camY}{\camX * 0.7}
\pgfmathsetmacro{\camZ}{0.8}

\newcommand{\camera}[4]{
  \begin{scope}[shift={#1},rotate around z=#2, every node/.style={text=#3}, every path/.style={draw=#3}]
    \draw (0,0,0) -- (\camZ, \camY/2.0,  -\camX/2.0);
    \draw (0,0,0) -- (\camZ, -\camY/2.0, -\camX/2.0);
    \draw (0,0,0) -- (\camZ, \camY/2.0,  \camX/2.0);
    \draw (0,0,0) -- (\camZ, -\camY/2.0, \camX/2.0);
    \draw (\camZ, \camY/2.0,  \camX/2.0) -- (\camZ, -\camY/2.0, \camX/2.0);
    \draw (\camZ, \camY/2.0,  -\camX/2.0) -- (\camZ, -\camY/2.0, -\camX/2.0);
    \draw (\camZ, \camY/2.0,  -\camX/2.0) -- (\camZ, \camY/2.0, \camX/2.0);
    \draw (\camZ, -\camY/2.0,  -\camX/2.0) -- (\camZ, -\camY/2.0, \camX/2.0);
    
    \node at (-0.9,0,0) {#4};
  \end{scope}
}

\node at (0.8,4,0) (cA) {};
\node at (0,1.5,0) (cL) {};
\node at (-0.5,0.4,0) (cGT) {};

\camera{(cA)}{192}{red!80}{\large$(\hat{q},\hat{t})_a$}
\camera{(cL)}{175}{blue!80}{\large$(\tilde{q},\tilde{t})_a$}
\camera{(cGT)}{168}{green!80}{\large$(q,t)_a$}

\draw[->, thick, bend left=45] (cA) to node[midway, right] (delta_node){\large$(\delta q, \delta t)_a$} (cL);

\draw[->, thick, dashed] (cA) .. controls (cL) .. (cGT);

\begin{scope}[shift={(2.9,2.2)}]
    \matrix (imgmatrix) [matrix of nodes,
          nodes={inner sep=0pt},
          column sep=0.0cm,
          row sep=0.3cm,
          anchor=west] {
    \node {Obs. Image}; & \node {Abs. Error}; \\
    \node[inner sep=0pt, name=img11]  {\includegraphics[width=2.5cm]{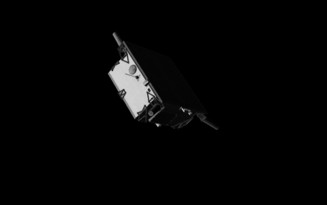}}; & \node[inner sep=0pt, name=img12] {\includegraphics[width=2.5cm]{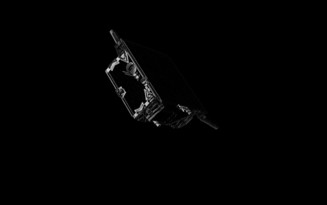}}; \\
    \node[inner sep=0pt, name=img21] {\includegraphics[width=2.5cm]{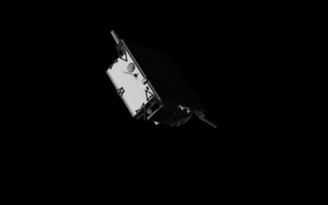}}; & \node[inner sep=0pt, name=img22] {\includegraphics[width=2.5cm]{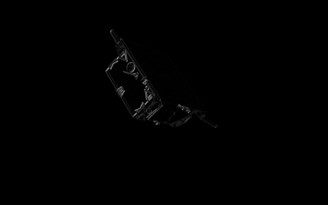}}; \\
    \node[inner sep=0pt, name=img31] {\includegraphics[width=2.5cm]{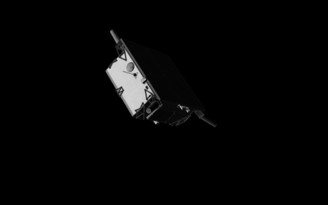}}; & \node[inner sep=0pt, name=img32] {\includegraphics[width=2.5cm]{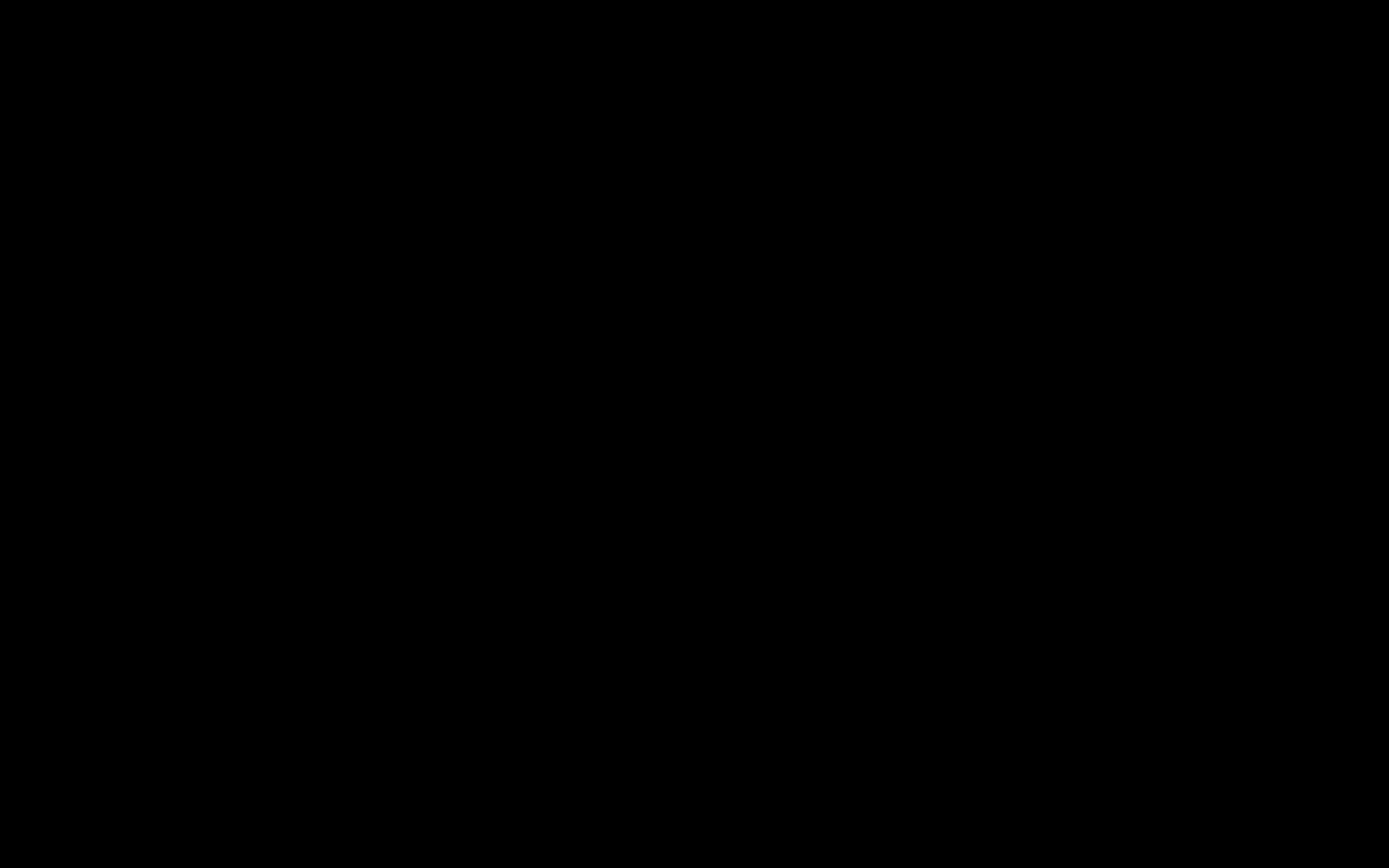}}; \\
    };
\end{scope}

\begin{pgfonlayer}{background}
    \node at (-4.6,2.0) {\includegraphics[width=5cm, trim={30cm, 30cm, 30cm, 30cm}]{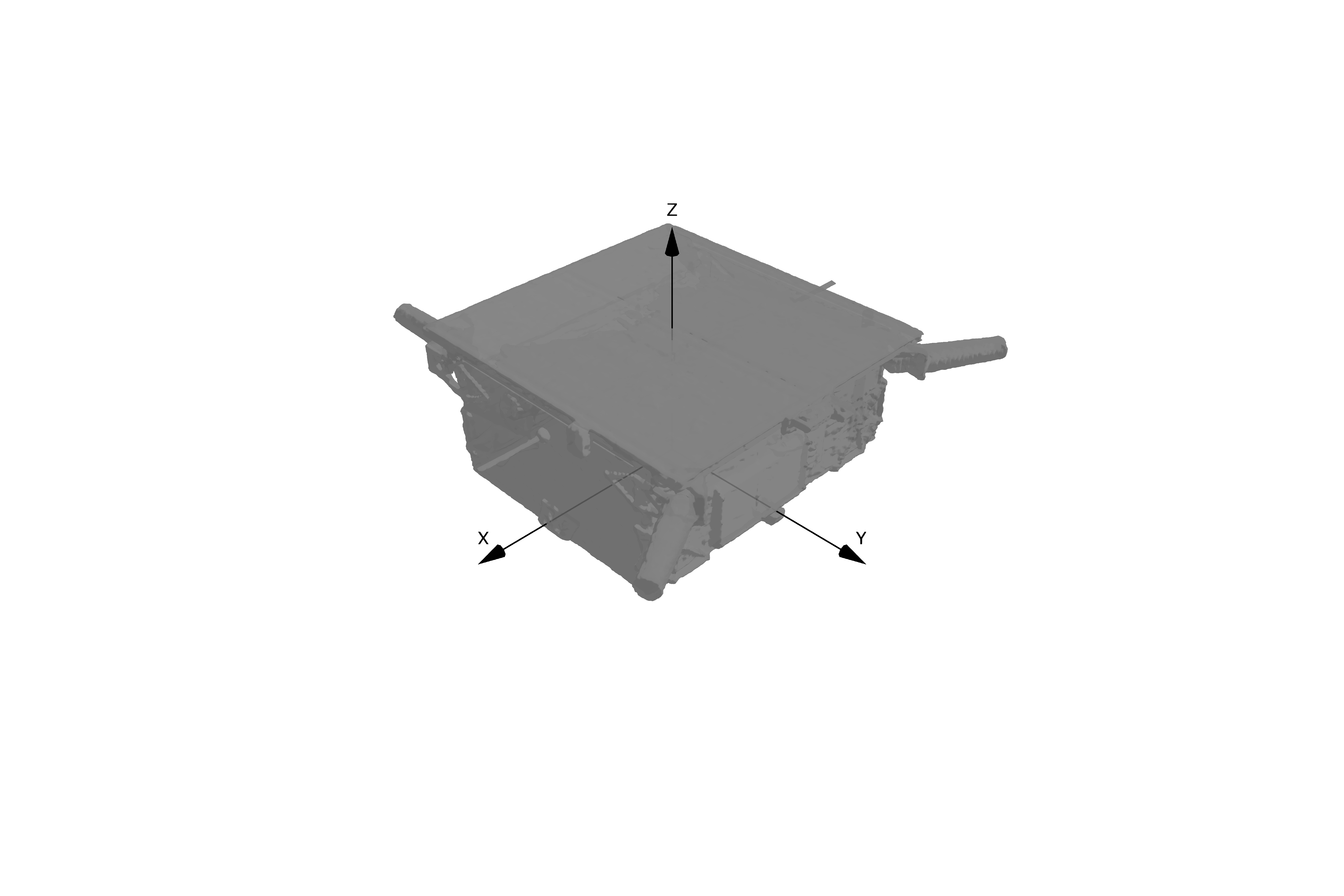}};
    
    \node[fill=red!25, draw=red!60, line width=0.25pt, rounded corners, inner sep=0.1cm, fit=(img11)(img12)] {};
    \node[fill=blue!25, draw=blue!60, line width=0.25pt, rounded corners, inner sep=0.1cm, fit=(img21)(img22)] {};
    \node[fill=green!25, draw=green!60, line width=0.25pt, rounded corners, inner sep=0.1cm, fit=(img31)(img32)] {};
\end{pgfonlayer}
\end{tikzpicture}
\vspace{-0.2cm}
\caption{\label{fig_pose_misalignment} Solving pose misalignment through learnable correction terms. The optimization goal of reducing the photometric error between the generated image and the observed one encourages the model to progressively refine the pose label.}
\end{figure}

To alleviate this uncertainty on the pose labels and therefore recover a sharper representation, we do not train the NeRF on the pseudo-labels directly but on refined pose labels, $\left\{(\tilde{q}_a, \tilde{\boldsymbol{t}}_a) \right\}_{a=1}^N$. 
As illustrated in \Cref{fig_pose_misalignment}, for each image $I_a$, the refined, learned, pose label is computed from the pseudo-label through a pose correction term, $(\delta q_a, \boldsymbol{\delta t}_a)$, learned along the NeRF training, \ie,
\begin{equation}
    \tilde{\boldsymbol{t}}_a = \hat{\boldsymbol{t}}_a + (\boldsymbol{\delta t})_a \quad \textnormal{and} \quad \tilde{q}_a =  (\delta q)_a \: \hat{q}_a
\end{equation}
where $\left\{(\boldsymbol{\delta t})_a\right\}_{a=1}^{N}$ are learnable parameters and $\left\{(\delta q)_a\right\}_{a=1}^{N}$ are correction terms formulated as quaternions, with scalar parts $\left\{(\delta q)^{\textnormal{0}}_a\right\}_{a=1}^{N}$ and vector parts $\left\{(\boldsymbol{\delta q)}^{\textnormal{vec}}_a\right\}_{a=1}^{N}$, \ie, 

\begin{equation}
    (\delta q)_a = 
    \begin{pmatrix}
        (\delta q)^{\textnormal{0}}_{a} \\ 
        (\boldsymbol{\delta q})^{\textnormal{vec}}_{a}
    \end{pmatrix}
\end{equation}
where $\left\{\boldsymbol{(\delta q)}^{\textnormal{vec}}_{a}\right\}_{a=1}^{N}$ are learnable parameters.

To ensure that the quaternion represents a valid rotation, the scalar component of the quaternion correction term, $(\delta q)^{\textnormal{0}}_{a}$, is not directly learned.
Instead, it is computed based on the vector part to enforce a unit-norm, \ie, 

\begin{equation}
      (\delta q)^{\textnormal{0}}_{a} = \sqrt{ 1 - \left| \left| \boldsymbol{(\delta q)}^{\textnormal{vec}}_a \right|\right|_2^2}.
\end{equation}

These pose correction terms can be seen as additional degrees of freedom granted to the model as they give the ability of fine-tuning the pose labels to improve the representation consistency, and in turn the rendering  sharpness. Indeed, as illustrated in \Cref{fig_pose_misalignment}, any misalignment of the pose labels results in an increased photometric error. Hence, the model, trained on the photometric loss, is encouraged to refine the pose labels as it contributes to reducing the photometric loss.

In practice, as explained in \Cref{sec_background}, neural radiance fields are not directly trained on images paired with their pose labels but rather on batches made of pixels, and their corresponding rays, sampled across all the training images to promote multi-view consistency and hence improve convergence.
Therefore, we do not apply the pose correction terms directly on the pose labels but on the rays as it is equivalent but easier to integrate in a NeRF.

The rays computed from the pseudo-pose labels ($\hat{\boldsymbol{c}}_{aij}$,$\hat{\boldsymbol{d}}_{aij}$), \ie, ray origin $\hat{\boldsymbol{c}}_{aij}$ and direction $\hat{\boldsymbol{d}}_{aij}$ as computed using the pseudo-pose labels ($\hat{q}_i$, $\hat{\boldsymbol{t}}_i$), are refined through the image-wise pose correction terms into fine-tuned rays ($\tilde{\boldsymbol{c}}_{aij}$, $\tilde{\boldsymbol{d}}_{aij}$), \ie, 
\begin{equation}
    \boldsymbol{\tilde{c}}_{aij} = \boldsymbol{\hat{c}}_{aij} + (\boldsymbol{\delta t})_a \quad \textnormal{and} \quad
    \begin{pmatrix}
        0 \\ 
        \boldsymbol{\tilde{d}}_{aij}
    \end{pmatrix} 
     = (\delta q)_a \otimes 
    \begin{pmatrix}
        0 \\ 
        \boldsymbol{\hat{d}}_{aij}
    \end{pmatrix} 
      \otimes (\delta q)_a^*
\end{equation}

Refining the pose labels does not significantly increase the training complexity as it is conducted during the training of the NeRF. 
In addition, it does not affect the inference complexity since the pose correction terms are not used to render novel views. 
The only hazard that arises from the pose fine-tuning is the risk of divergence that could happen if the pose labels move away from the true camera poses during the first training steps. 
However, by imposing lower learning rates on the pose correction terms, this risk is reduced to a negligible level and has never been observed, \ie,

\begin{equation}
    l_{r}^{(\delta t)} = \beta^{(\delta t)} \: l_{r}   \quad \textnormal{and} \quad l_{r}^{(\delta q)} = \beta^{(\delta q)} \: l_{r} ,
\end{equation}
where $l_{r}^{(\delta t)}$ and $l_{r}^{(\delta q)}$ are the learning rates for the position and orientation correction terms, $\beta^{(\delta t)}$ and $\beta^{(\delta q)}$ are factors lower or equal to 0.01 to ensure convergence, and $l_{r}$ is the NeRF learning rate.

\setlength{\tabcolsep}{0pt}
\renewcommand{\arraystretch}{1}

\begin{table}[b!]
\centering
\caption{\small\label{tab_datasets} Overview of the 3 validation sets}
\begin{NiceTabular}{|@{}>{\centering\arraybackslash}m{1.9cm}@{}*{3}{|c@{}}|}[
  cell-space-top-limit=0pt,
  cell-space-bottom-limit=0pt
]
\hline
\centering Set & \textit{Synthetic} & \textit{Lightbox} & \textit{Sunlamp} \\
\hline
\centering Dataset & SHIRT~\cite{park2023adaptive} & SHIRT~\cite{park2023adaptive} & SPEED+~\cite{park2022speedplus}\\
\hline
\centering Target & Simplified CAD model & \multicolumn{2}{c|}{Realistic mockup (HIL)} \\
\hline
\centering Illumination & Synthetic & Diffuse (Earth Albedo) & Direct (Sun) \\
\hline
\centering Pose distr. & \multicolumn{2}{c|}{Uniform sampling over a single trajectory} & Random $\mathrm{SE}(3)$ Sampling \\
\hline
\centering Pose labels & Ground-Truth & \multicolumn{2}{c|}{Realistic Pseudo-labels} \\
\hline
\centering \# images & 2371 & 2371 & 2791 \\
\hline
\adjustbox{valign=m}{Examples} &
\raisebox{-1.1cm}{\smash{\includegraphics[width=0.29\linewidth]{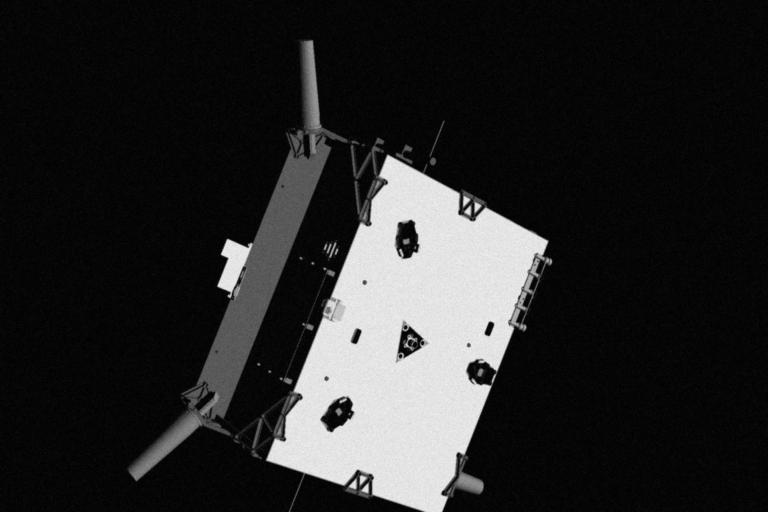}}}\rule{0pt}{2.1cm} &
\raisebox{-1.1cm}{\smash{\includegraphics[width=0.29\linewidth]{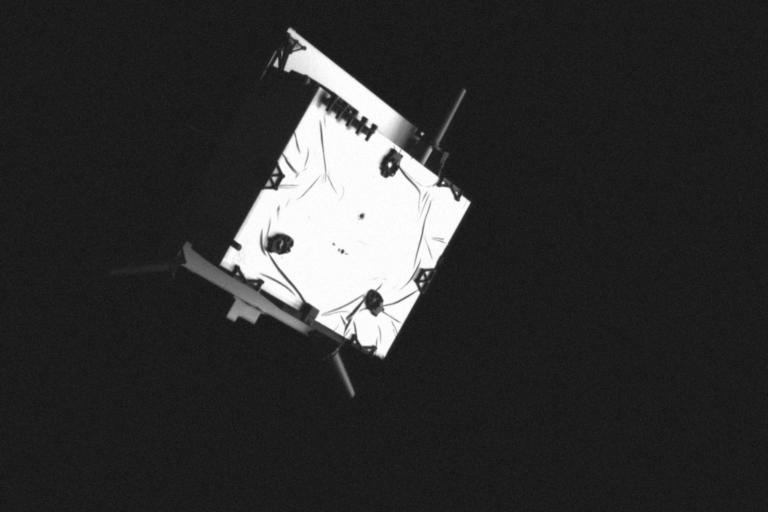}}}\rule{0pt}{2.1cm} &
\raisebox{-1.1cm}{\smash{\includegraphics[width=0.29\linewidth]{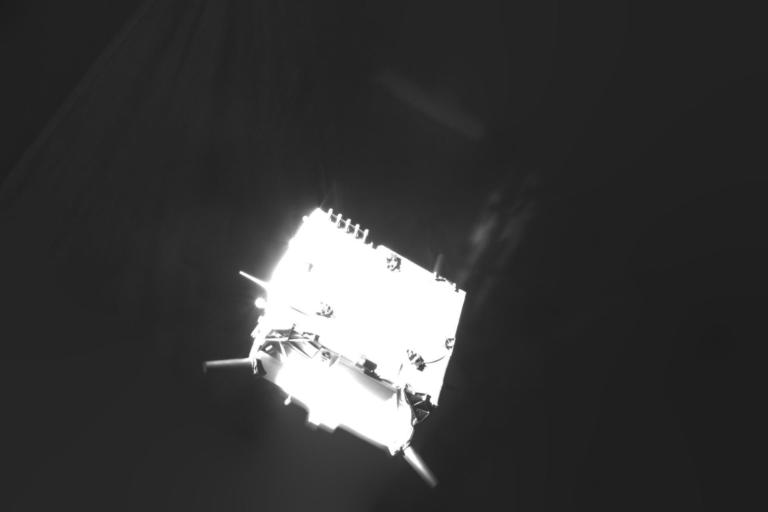}}}\rule{0pt}{2.1cm} \\
\hline
\end{NiceTabular}
\end{table}

\section{Datasets}
\label{sec_datasets}

This section describes the datasets on which we conducted experiments to demonstrate the effectiveness of the proposed method. Those sets are summarized in \Cref{tab_datasets}. Three sets were selected from the SPEED+~\cite{park2022speedplus} and SHIRT~\cite{park2023adaptive} datasets. Both datasets contain images depicting the TANGO satellite from the PRISMA mission~\cite{gill2007autonomous}, taken with the same camera parameters. Each dataset contains both synthetic and Hardware-In-the-Loop (HIL) images along their respective pose labels. While SPEED+ contains images generated independently from pose labels randomly sampled in SE(3), SHIRT consists in two trajectories (ROE), where the camera is approaching the target spacecraft. For each time step of the trajectory, a synthetic and a HIL image are generated using the same pose label.

The synthetic images were rendered by an OpenGL-based tool which synthesizes the images given the corresponding pose labels. The OpenGL tool, as it relies on rasterization, hardly renders the illumination conditions encountered in orbit. Furthermore, it relies on a simplified CAD model of the target spacecraft, which lacks of textural details, such as the Multi-Layer Insulation (MLI). 

The Hardware-In-the-Loop (HIL) images, also dubbed as \textit{real} images, were acquired in the Testbed for Rendezvous and Optical Navigation (TRON)~\cite{park2021robotic} facility at Stanford University. This testbed replicates the illumination conditions encountered in orbit on a realistic mock-up of the target satellite. The direct illumination from the Sun is emulated through a metal halide arc lamp while diffuse illumination coming from the Earth albedo is simulated through light boxes. The pose labels associated to the acquired images are estimated through infrared markers placed on the robot arms that moves the camera and as well as the target. Hence, the pose labels suffer from some uncertainty as they would with real in-orbit images annotated on ground.

In the following, \textit{Synthetic} denotes the set of synthetic images corresponding to the second trajectory (ROE) of SHIRT. The first ROE is not considered here as two of the target faces never appear in the images, and could therefore not lead to a consistent 3D reconstruction. Similarly, \textit{Lightbox} consists in the HIL images of the second ROE of SHIRT. As a result, both sets contain images taken from the same viewpoints, apart from the slight calibration error that is inherent to the HIL setup. Additional differences between both sets are the diffuse illumination that emulates the earth albedo as well as an improved realism achieved through the target mock-up. The key differences between both sets are therefore the realistic illumination conditions as well as the pose uncertainty, which are the specific challenges addressed by our method. Finally, \textit{Sunlamp}, a set from SPEED+, consists in images taken under even more challenging illumination conditions. Their acquisition setup emulates the direct illumination from the Sun on a specular target. The \textit{Sunlamp} images are therefore of a poorer quality and are characterized by under as well as over-exposure, including sun flares.  

The 3 sets have different purposes. \textit{Lightbox} is representative of a real use case and allows us to validate the ability of our method to handle varying illumination conditions as well as pose uncertainty. Similarly, \textit{Sunlamp} is representative of the most challenging illumination conditions that can be encountered on-orbit. By demonstrating the ability of our method in handling such challenging illumination conditions, we show that our method is robust to a wide range of illumination scenarios. Finally, even if the \textit{Synthetic} set is over-simplified compared to the real ones, it offers a controllable environment to perform ablation studies on the pose finetuning abilities of our method. This allows us to demonstrate the ability of our method to cope with a wide range of noise on the pseudo-pose labels.

\section{Experiments}
\label{sec_experiments}

This section describes and analyzes the experiments conducted to evaluate the proposed 3D reconstruction method. \Cref{sec_experiments_setup} describes our validation setup while \Cref{sec_experiments_main} evaluates the benefits brought by our method compared to a standard, off-the-shelf, NeRF. \Cref{sec_experiments_nimages} assesses our method resilience to a limited number of images. An ablation study, conducted on the proposed pose correction mechanism (see \Cref{sec_method_pose_finetuning}), is carried out in \Cref{sec_experiments_finetuning}.

\subsection{Validation Setup}
\label{sec_experiments_setup}
This section describes the training and evaluation setup used to assess our method. In the following, "baseline" refers to a model trained without appearance embeddings nor pose correction terms, as opposed to our model.

\subsubsection*{Training}
    Unless otherwise mentioned, each model evaluated in this paper is trained on 400 images randomly sampled in the considered set. All images are preprocessed to remove the background using foreground segmentation masks. Each model, along with the 400 appearance embeddings and pose correction terms, is trained for 30,000 optimization steps of 4096 rays per batch. The training takes 30 minutes on an NVIDIA L40s. 

\subsubsection*{Evaluation}

    Since the 3D model of TANGO is not available, evaluating directly the accuracy of the learned 3D representation is infeasible. Instead, we evaluate the 2D image reconstruction quality as a proxy of the 3D accuracy. For this purpose, we randomly picked 200 pose labels, along the corresponding images, from a realistic set and evaluate the ability of our model in reconstructing the images from the pose labels. The \textit{Lightbox} set of SPEED+~\cite{park2022speedplus} was chosen for this purpose as it contains realistic images of the same target under viewpoints and illumination conditions that are different from the ones encountered in the different training sets.

    This photometric-based evaluation suffers from an intrinsic ambiguity. Indeed, the same object viewed under the same pose corresponds to multiple images, depending on the illumination conditions. We solve this ambiguity through a widely accepted practice in the novel view synthesis community~\cite{martin2021nerf}. For each generated image, we first train an appearance embedding using half of the reference image pixels to implicitly learn the illumination conditions. This embedding is then used to generate the full image using illumination conditions aligned with the reference image, thereby enabling a valid evaluation of the reconstruction quality. 

    Each generated image is then compared to the reference one through qualitative as well as quantitative assessment. For each pair of images, we compute the PSNR,  SSIM~\cite{wang2004image} and JNB score~\cite{ferzli2009no}. The JNB score, \ie Just Noticeable Blur, is a no-reference metric which quantifies the amount of blur in an image, and hence, the sharpness of its content. Unless otherwise mentioned the values reported in this paper correspond to the metric averaged on the 200 test images.

\setlength{\tabcolsep}{0pt} 
\renewcommand{\arraystretch}{0} 
\begin{table}[t]
\centering
\caption{\small\label{tab_validation_roe2_synthetic} Qualitative assessment of our reconstruction method applied on \textit{Synthetic} images, compared to the baseline reconstruction. Both methods perform well on synthetic data, although the baseline struggles to handle illumination variations}
\begin{NiceTabular}{@{}>{\centering\arraybackslash}m{1.7cm}@{}*{5}{c@{}}}[hvlines]
\centering \raisebox{22pt}[0pt][0pt]{Reference} & 
    \includegraphics[width=0.18\textwidth]{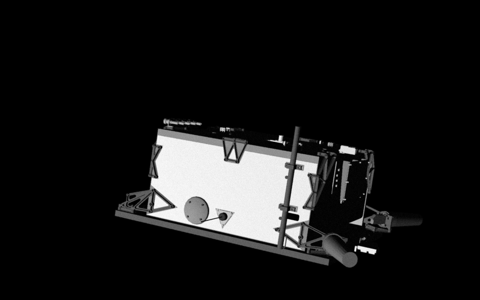} &
    \includegraphics[width=0.18\textwidth]{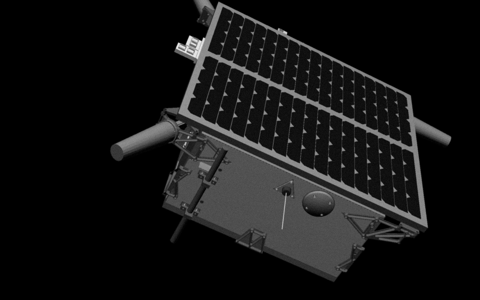} &
    \includegraphics[width=0.18\textwidth]{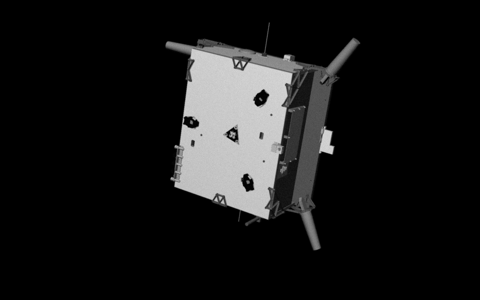} &
    \includegraphics[width=0.18\textwidth]{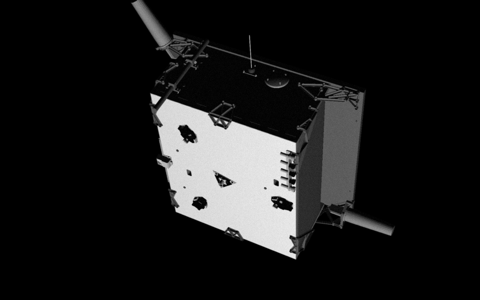} &
    \includegraphics[width=0.18\textwidth]{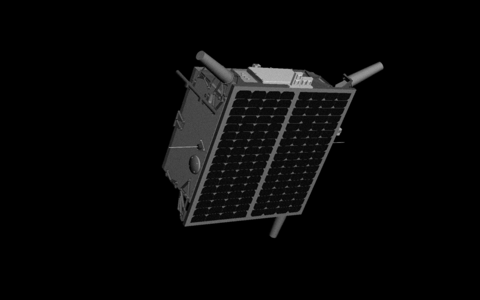} \\
\centering \raisebox{22pt}[0pt][0pt]{Baseline} & 
    \includegraphics[width=0.18\textwidth]{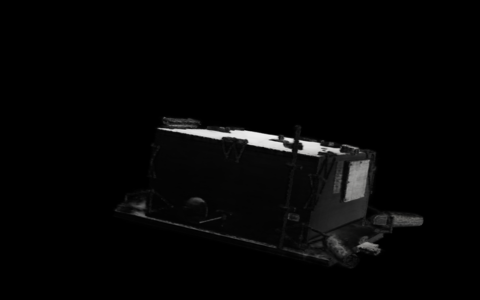} &
    \includegraphics[width=0.18\textwidth]{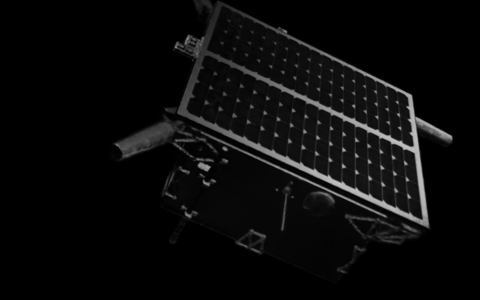} &
    \includegraphics[width=0.18\textwidth]{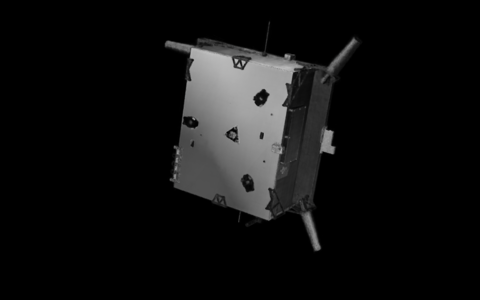} &
    \includegraphics[width=0.18\textwidth]{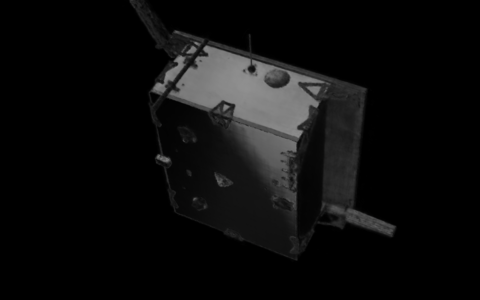} &
    \includegraphics[width=0.18\textwidth]{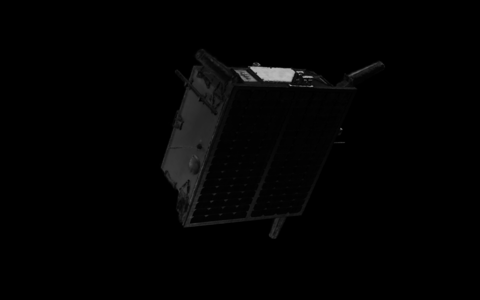} \\
\centering \raisebox{19pt}[0pt][0pt]{\shortstack{Ours}} & 
    \includegraphics[width=0.18\textwidth]{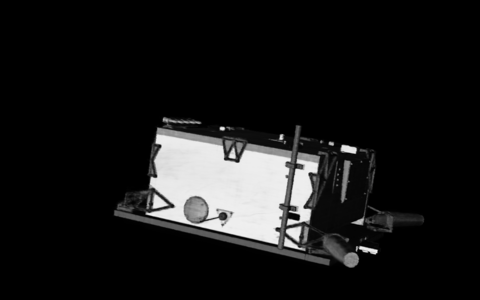} &
    \includegraphics[width=0.18\textwidth]{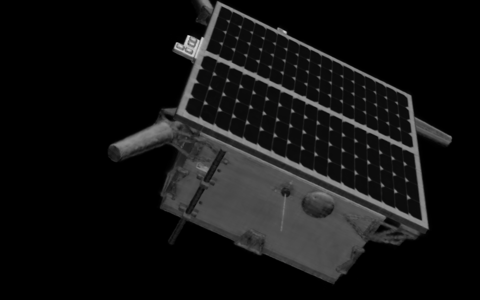} &
    \includegraphics[width=0.18\textwidth]{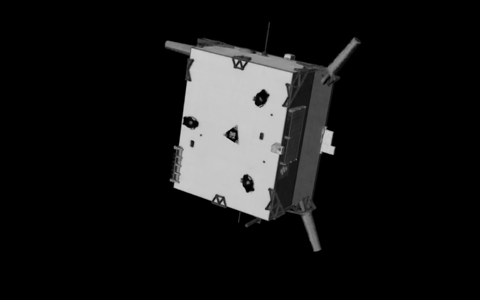} &
    \includegraphics[width=0.18\textwidth]{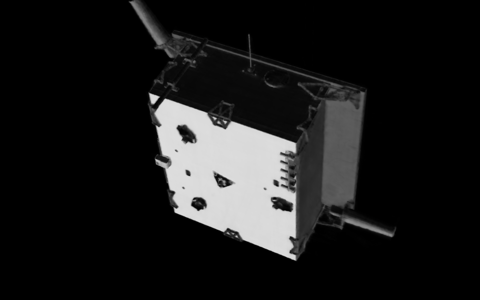} &
    \includegraphics[width=0.18\textwidth]{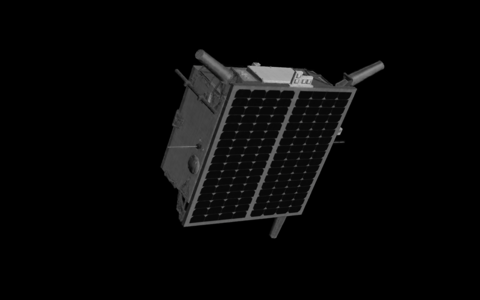} \\
\end{NiceTabular}
\end{table}

\setlength{\tabcolsep}{0pt} 
\renewcommand{\arraystretch}{0} 
\begin{table}[t]
\centering
\caption{\small\label{tab_validation_roe2_lightbox} Qualitative assessment of our reconstruction method, with and without pose finetuning, applied on \textit{Lightbox} images, compared to the baseline reconstruction. While the baseline struggles to handle varying illumination, our method achieves it. In addition, pose finetuning enables the learning of a sharper representation as it reduces the 3D uncertainty.}
\begin{NiceTabular}{|@{}m{0.6cm}|m{1.1cm}|*{5}{c}@{}|}
\hline
\multicolumn{2}{m{1.7cm}}{ \centering\raisebox{21pt}[0pt][0pt]{Reference} } &
    \includegraphics[width=0.18\textwidth]{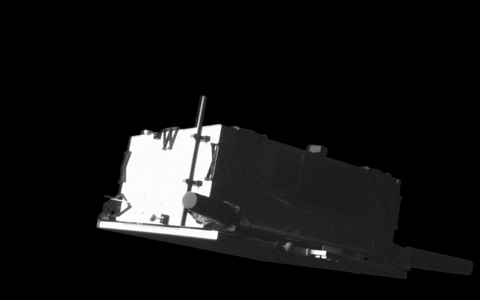} &
    \includegraphics[width=0.18\textwidth]{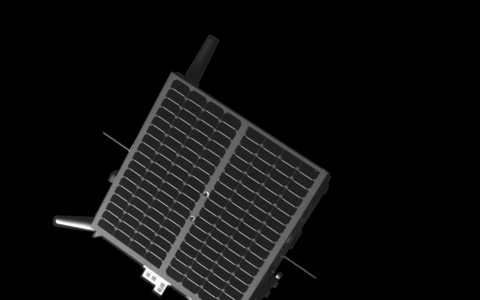} &
    \includegraphics[width=0.18\textwidth]{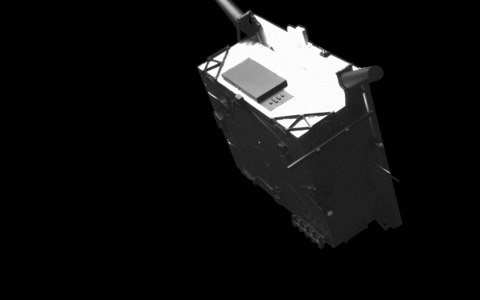} &
    \includegraphics[width=0.18\textwidth]{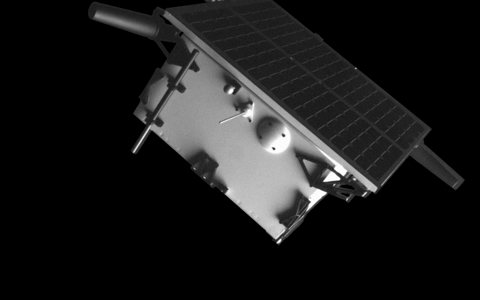} &
    \includegraphics[width=0.18\textwidth]{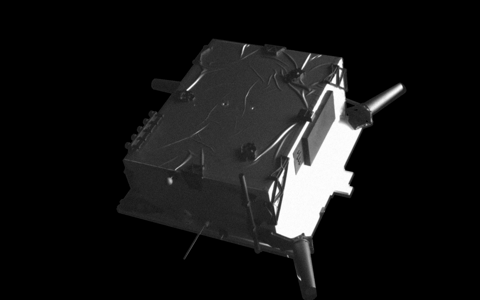} \\
\hline
\multicolumn{2}{m{1.7cm}}{\centering\raisebox{21pt}[0pt][0pt]{ Baseline}} &
    \includegraphics[width=0.18\textwidth]{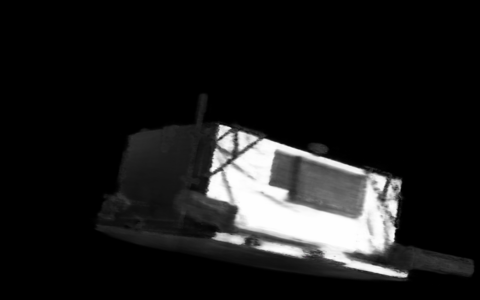} &
    \includegraphics[width=0.18\textwidth]{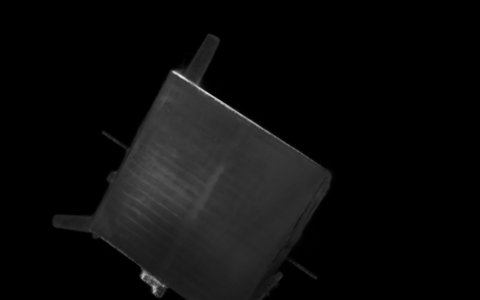} &
    \includegraphics[width=0.18\textwidth]{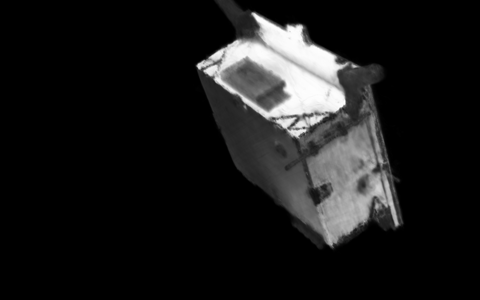} &
    \includegraphics[width=0.18\textwidth]{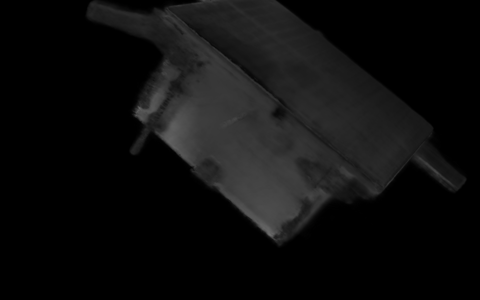} &
    \includegraphics[width=0.18\textwidth]{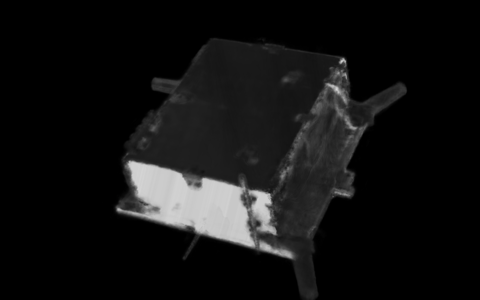} \\
\hline
\multirow{2}{*}{\hspace{0.2cm}\rotatebox{90}{\hspace{0.4cm} Ours \hspace{-0.4cm}}} & \centering \raisebox{14pt}[0pt][0pt]{\shortstack{\small W/o. \\ pose \\ corr.}}  &
    \includegraphics[width=0.18\textwidth]{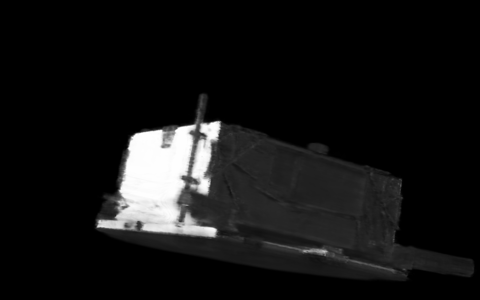} &
    \includegraphics[width=0.18\textwidth]{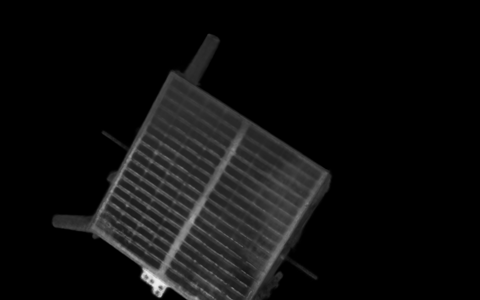} &
    \includegraphics[width=0.18\textwidth]{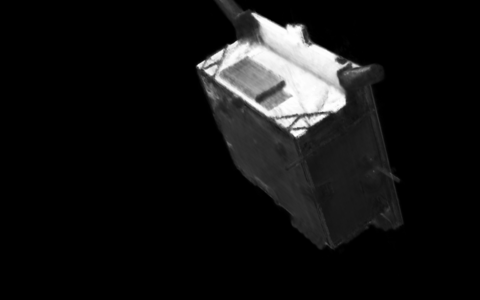} &
    \includegraphics[width=0.18\textwidth]{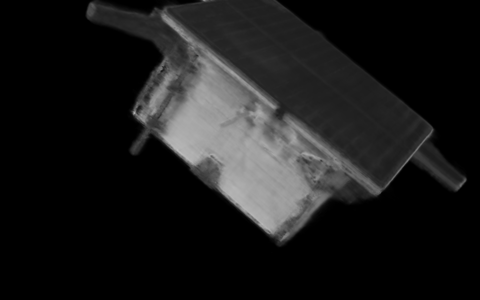} &
    \includegraphics[width=0.18\textwidth]{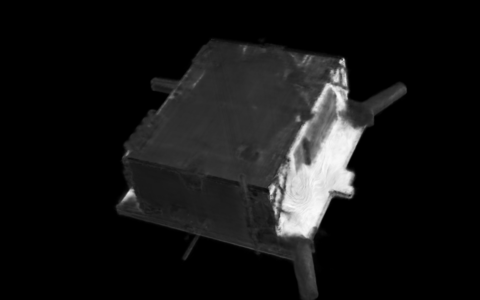} \\
\cline{2-7}
& \centering \raisebox{14pt}[0pt][0pt]{\shortstack{\small W. \\ pose \\ corr.}}  &
    \includegraphics[width=0.18\textwidth]{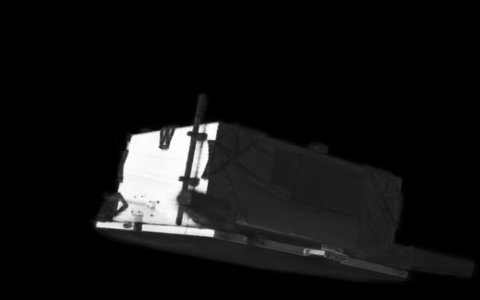} &
    \includegraphics[width=0.18\textwidth]{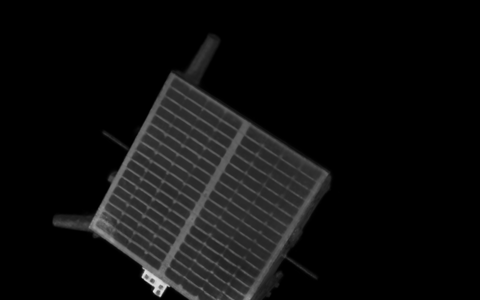} &
    \includegraphics[width=0.18\textwidth]{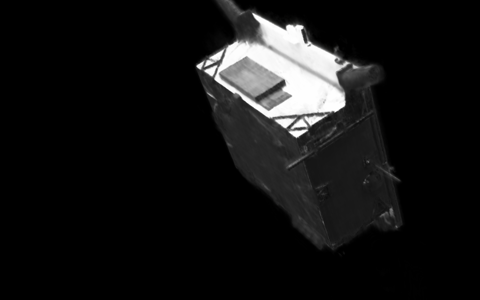} &
    \includegraphics[width=0.18\textwidth]{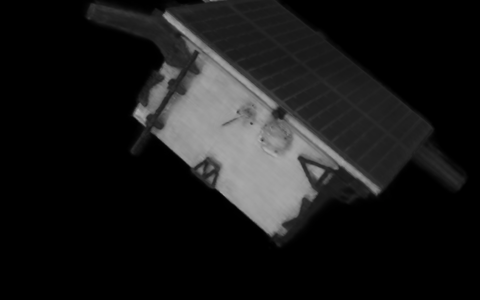} &
    \includegraphics[width=0.18\textwidth]{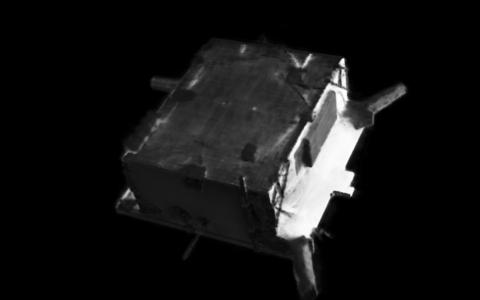} \\
\hline
\end{NiceTabular}
\end{table}

\subsection{Validation}
\label{sec_experiments_main}

This section compares the image reconstruction quality of our method against the baseline. \Cref{tab_validation_roe2_synthetic} shows images generated from identical viewpoints for both models trained on synthetic images with accurate pose labels. Both approaches successfully capture the 3D geometry of the target spacecraft. However, while our method accurately handles illumination conditions, the baseline exhibits artifacts such as shadows in the rendered images. This improvement stems from our model’s ability to decouple scene geometry from image-specific illumination.

Across all image pairs, reconstructions produced by our method consistently achieve higher PSNR and SSIM values than those generated by the baseline. Over 200 images, our method reaches an average PSNR of 30.3 dB, compared to 22.0 dB for the baseline, confirming a substantial quality improvement. The same trend holds for SSIM: 0.948 for our method versus 0.927 for the baseline. A key factor in this illumination consistency is the use of appearance embeddings, which align lighting conditions between reference and generated images, thereby improving perceptual quality.

\setlength{\tabcolsep}{0pt}
\renewcommand{\arraystretch}{3.5}
\begin{table}[t]
\centering
\caption{\small\label{tab_pose_finetuning_roe2_lightbox} Enlargement of the images depicted in \Cref{tab_validation_roe2_lightbox} to highlight their high-frequency content. Finetuning the pose labels results in a sharper representation, which finally lowers the level of blur on the generated images.}
\begin{NiceTabular}{@{}m{1.5cm}@{}*{5}{@{}>{\centering\arraybackslash}m{3cm}@{}}}[hvlines]
\Block{2-1}{Reference} & 
    \Block{2-1}{\imagecellnew{3cm}{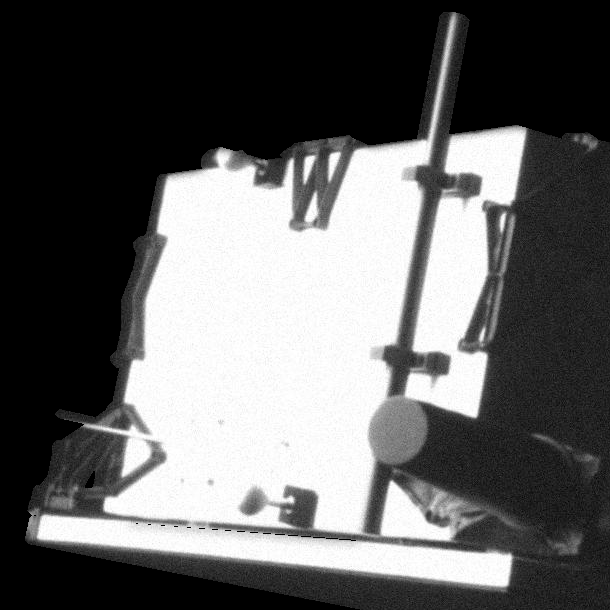}} &
    \Block{2-1}{\imagecellnew{3cm}{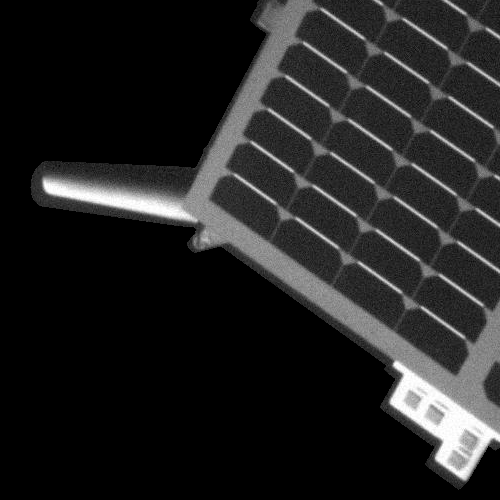}} &
    \Block{2-1}{\imagecellnew{3cm}{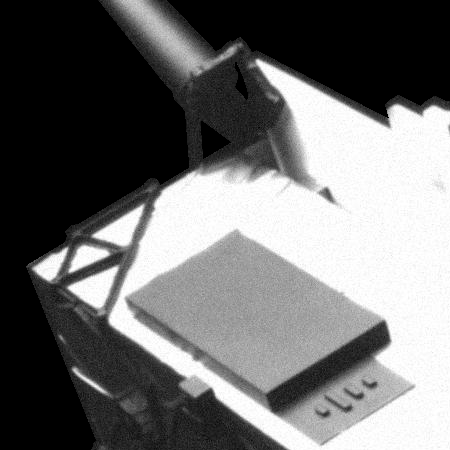}} &
    \Block{2-1}{\imagecellnew{3cm}{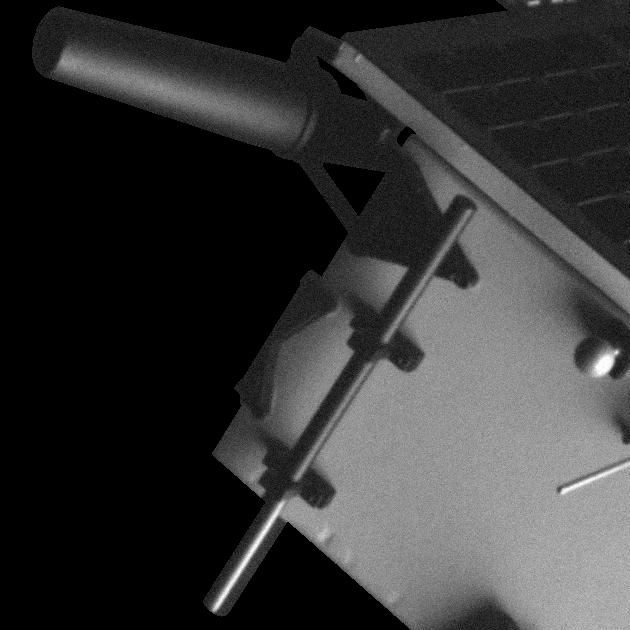}} &
    \Block{2-1}{\imagecellnew{3cm}{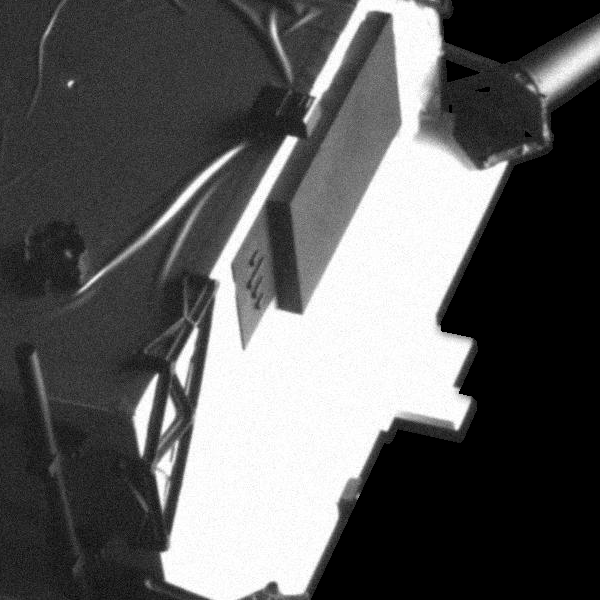}} \\
& & & & & \\
\Block{2-1}{W/o. pose corr.} & 
    \Block{2-1}{\imagecellnew{3cm}{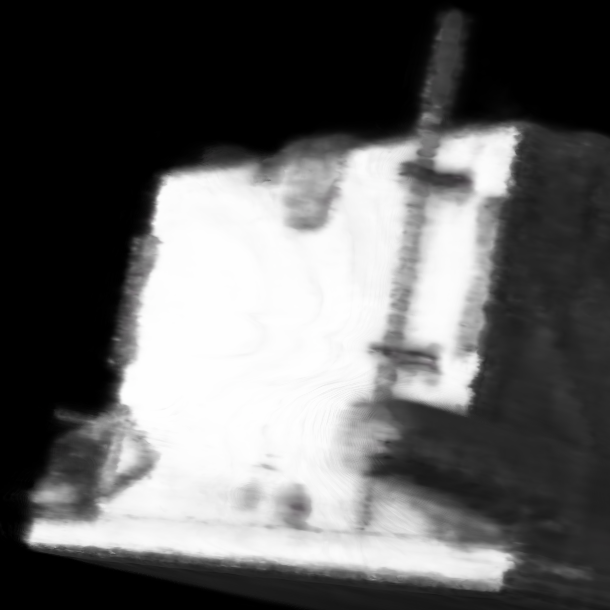}} &
    \Block{2-1}{\imagecellnew{3cm}{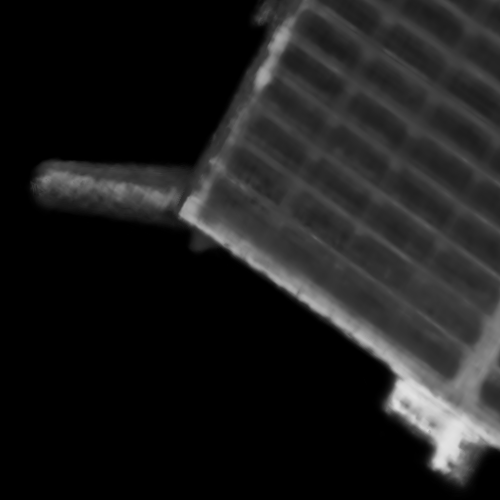}} &
    \Block{2-1}{\imagecellnew{3cm}{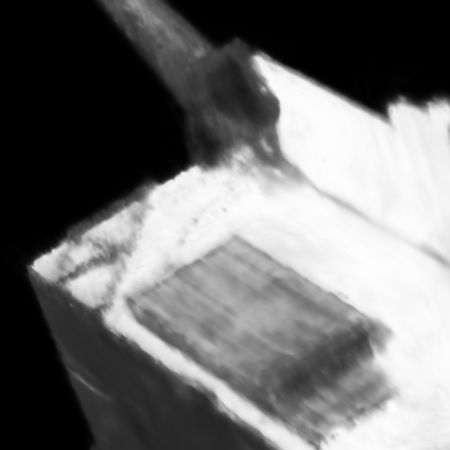}} &
    \Block{2-1}{\imagecellnew{3cm}{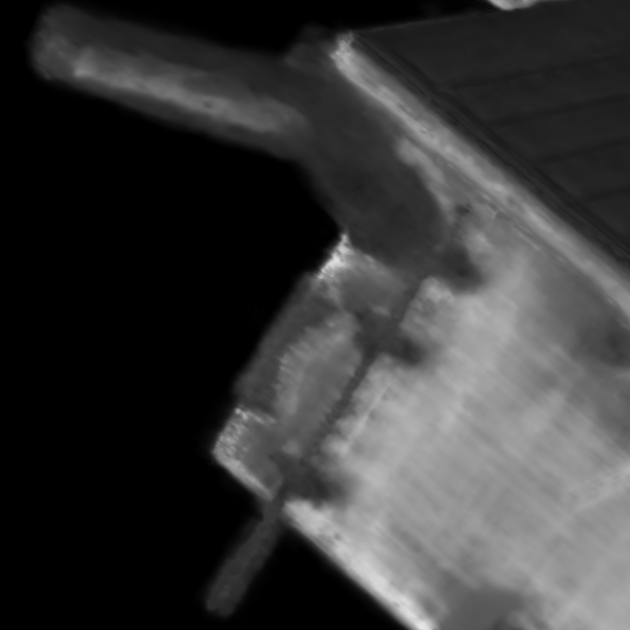}} &
    \Block{2-1}{\imagecellnew{3cm}{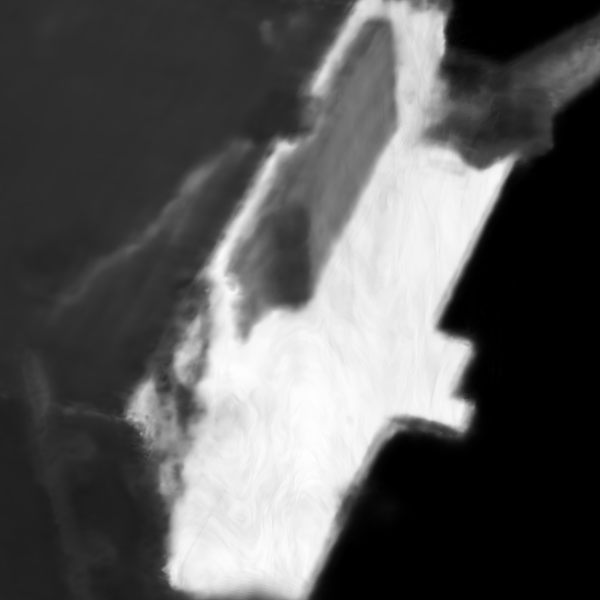}} \\
& & & & & \\
\Block{2-1}{W. pose corr.} & 
    \Block{2-1}{\imagecellnew{3cm}{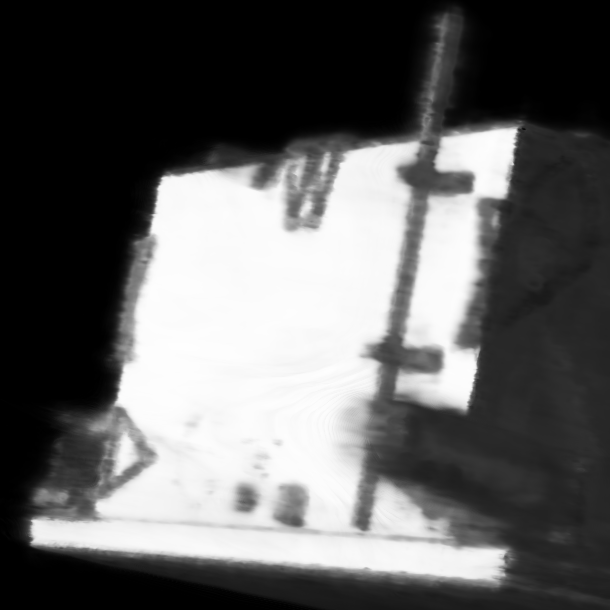}} &
    \Block{2-1}{\imagecellnew{3cm}{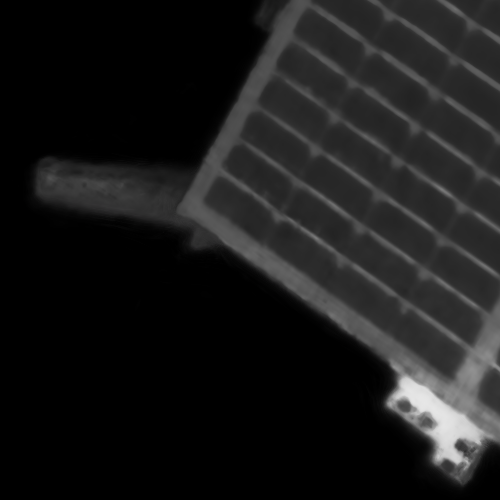}} &
    \Block{2-1}{\imagecellnew{3cm}{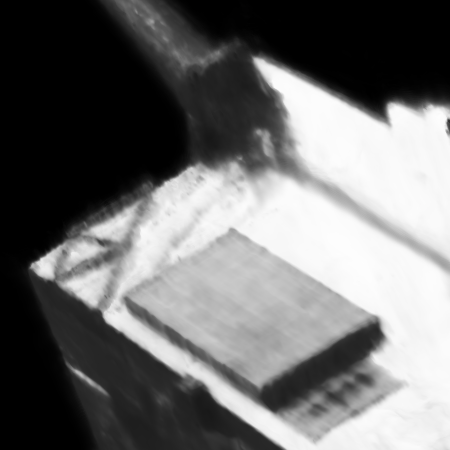}} &
    \Block{2-1}{\imagecellnew{3cm}{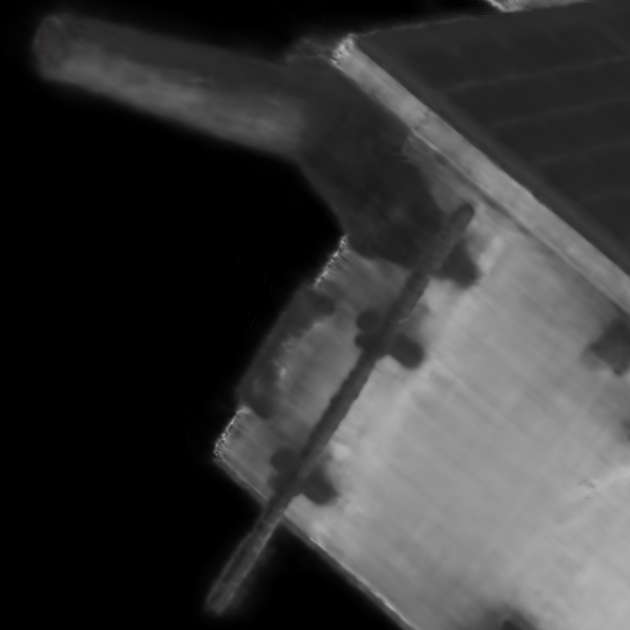}} &
    \Block{2-1}{\imagecellnew{3cm}{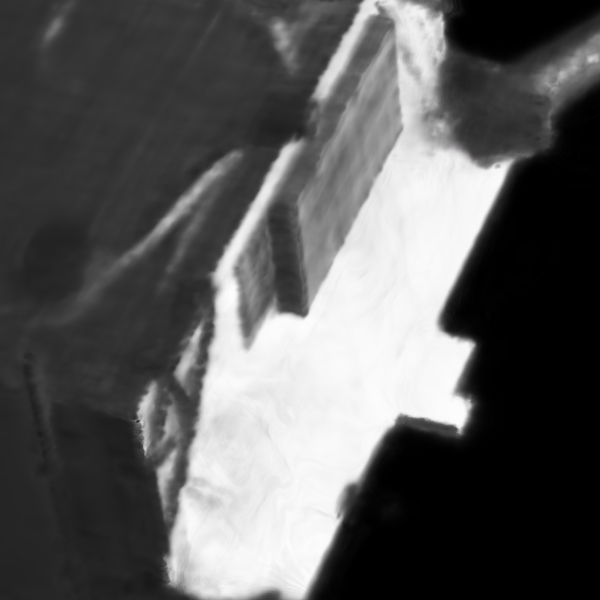}} \\
& & & & & \\
\end{NiceTabular}
\end{table}

\setlength{\tabcolsep}{0pt} 
\renewcommand{\arraystretch}{0} 
\begin{table}[t]
\centering
\caption{\small\label{tab_validation_sunlamp} Qualitative assessment of our reconstruction method, with and without pose correction, applied on \textit{Sunlamp} images, compared to the baseline reconstruction. The baseline fails in capturing the illumination conditions while our method is able to cope the challenging illumination conditions encountered in \textit{Sunlamp}. Furthermore, correcting the pose labels improves the representation's sharpness.}
\begin{NiceTabular}{|@{}m{0.6cm}|m{1.1cm}|*{5}{c}@{}|}
\hline
\multicolumn{2}{m{1.7cm}}{\centering\raisebox{21pt}[0pt][0pt]{ Baseline}} &
    \includegraphics[width=0.18\textwidth]{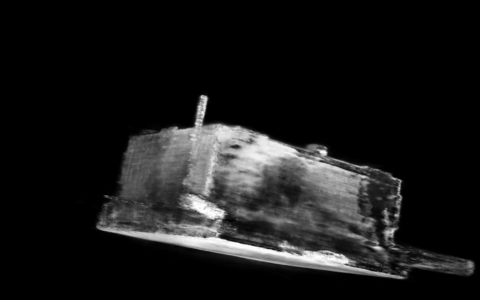} &
    \includegraphics[width=0.18\textwidth]{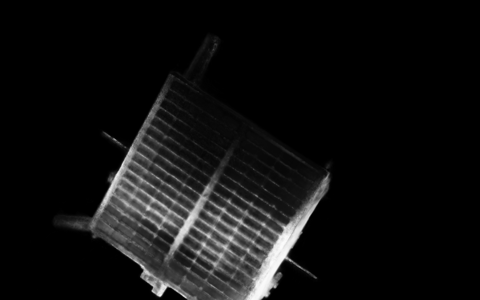} &
    \includegraphics[width=0.18\textwidth]{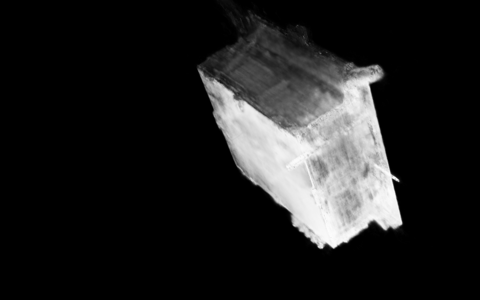} &
    \includegraphics[width=0.18\textwidth]{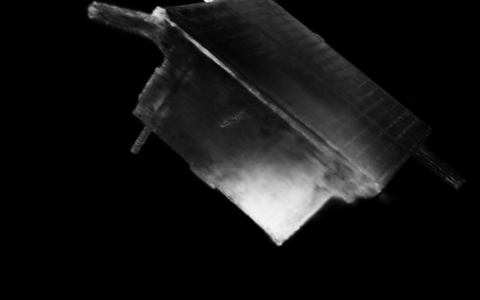} &
    \includegraphics[width=0.18\textwidth]{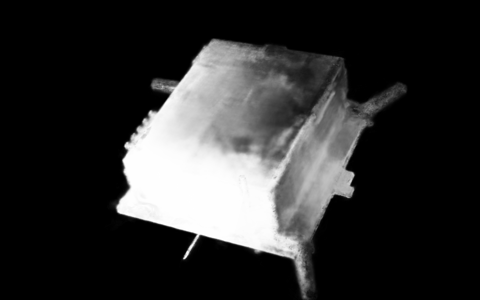} \\
\hline
\multirow{2}{*}{\hspace{0.2cm}\rotatebox{90}{\hspace{0.4cm} Ours \hspace{-0.4cm}}} & \centering \raisebox{14pt}[0pt][0pt]{\shortstack{\small W/o. \\ pose \\ corr.}}  &
    \includegraphics[width=0.18\textwidth]{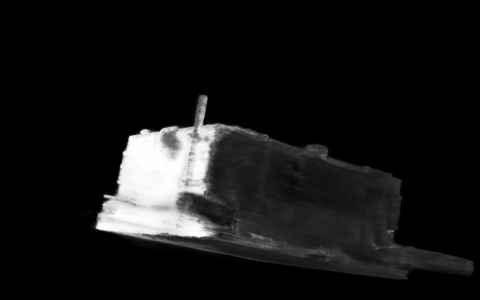} &
    \includegraphics[width=0.18\textwidth]{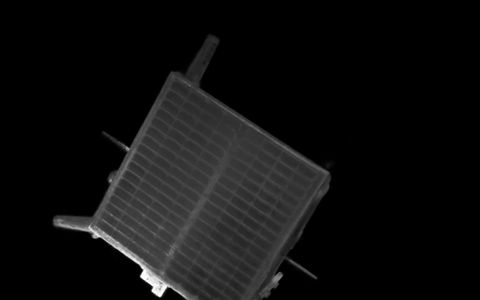} &
    \includegraphics[width=0.18\textwidth]{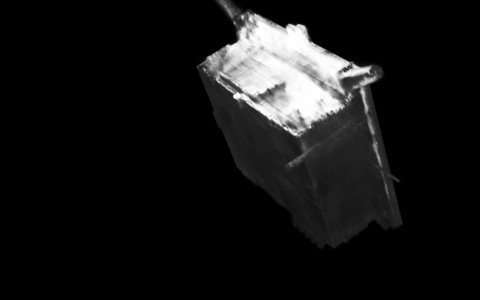} &
    \includegraphics[width=0.18\textwidth]{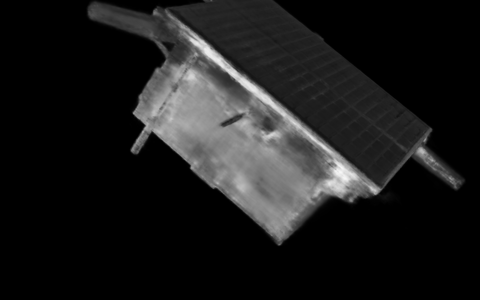} &
    \includegraphics[width=0.18\textwidth]{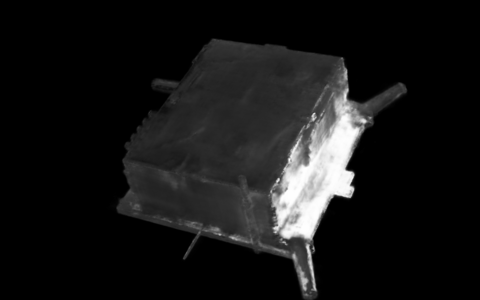} \\
\cline{2-7}
& \centering \raisebox{14pt}[0pt][0pt]{\shortstack{\small W. \\ pose \\ corr.}}  &
    \includegraphics[width=0.18\textwidth]{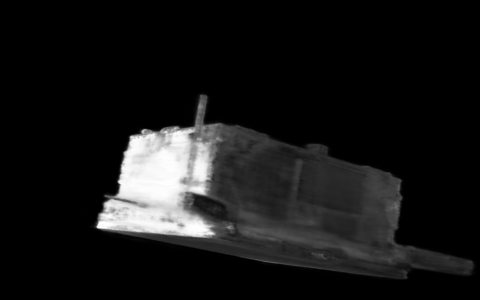} &
    \includegraphics[width=0.18\textwidth]{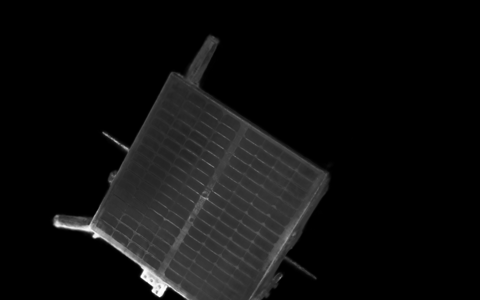} &
    \includegraphics[width=0.18\textwidth]{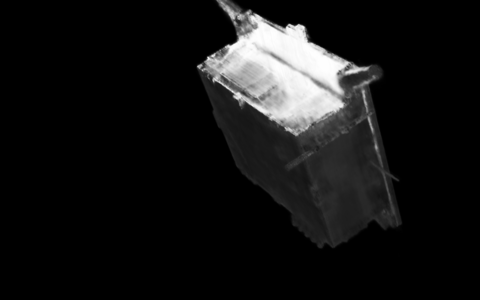} &
    \includegraphics[width=0.18\textwidth]{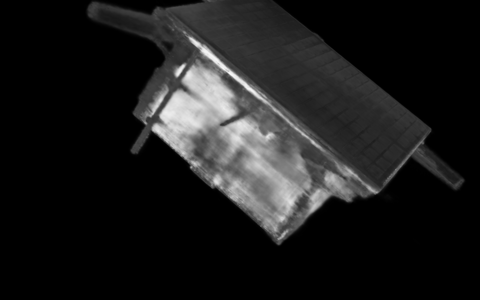} &
    \includegraphics[width=0.18\textwidth]{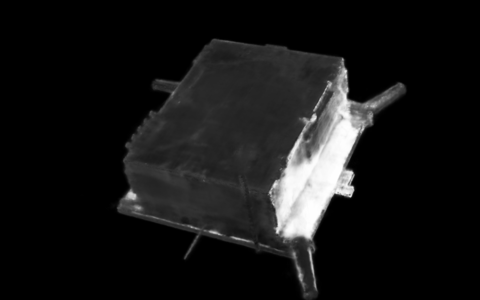} \\
\hline
\end{NiceTabular}
\end{table}

\Cref{tab_validation_roe2_lightbox} presents results on the \textit{Lightbox} dataset, which better reflects real mission conditions, including orbital lighting and pose uncertainty. Our method successfully reconstructs both geometry and appearance, whereas the baseline struggles with illumination, producing averaged lighting effects. Incorporating appearance embeddings improves PSNR from 24.35 dB to 30.90 dB and SSIM from 0.944 to 0.952.

Finetuning the pose labels further enhances representation sharpness and image detail. For example, solar panels reconstructed by the baseline appear textureless, while our method recovers structural details and even individual solar cells. When combined with pose correction, these details become sharper, as illustrated in \Cref{tab_pose_finetuning_roe2_lightbox}, which magnifies high-frequency content. Features such as solar cells, torque rods, antennas, and MLI wrinkles are significantly sharper when pose correction is applied. These capabilities are analyzed in \Cref{sec_experiments_finetuning}. Using pose correction, generated images achieve an average PSNR of 30.54 dB, SSIM of 0.950, and JNB score of 0.0333. Although PSNR and SSIM slightly decrease, the JNB score improves by 5.8\%, indicating a notable sharpness gain.

\setlength{\tabcolsep}{0pt} 
\renewcommand{\arraystretch}{0} 
\begin{table}[b!]
\centering
\caption{\small\label{tab_nimages_roe2_lightbox} Qualitative assessment of our reconstruction method applied on \textit{Lightbox} images, compared to the baseline reconstruction, for different number of training images. Our method is more resilient to scarce training sets.}
\begin{NiceTabular}{m{1.5cm}@{}*{6}{c@{}}}[hvlines]
\vspace{0.1cm} \, \# images \vspace{0.2cm} & 25 & 50 & 100 & 200 & 400& \small Reference \\
\vspace{-0.6cm}\centering \small Baseline \vspace{0.6cm} & \includegraphics[width=0.15\textwidth]{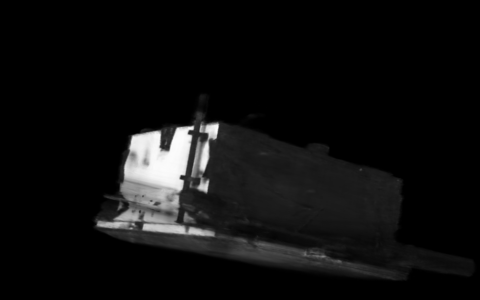} &
    \includegraphics[width=0.15\textwidth]{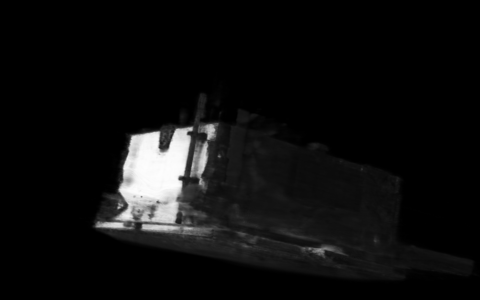} &
    \includegraphics[width=0.15\textwidth]{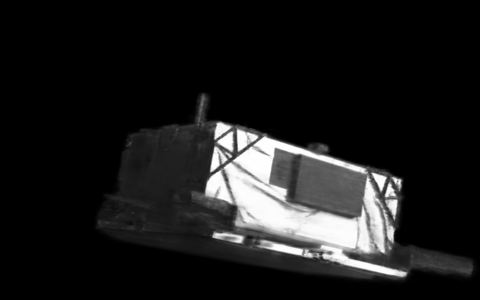} &
    \includegraphics[width=0.15\textwidth]{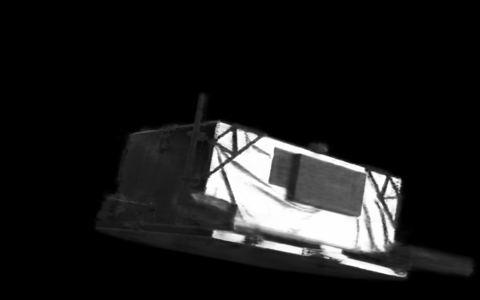} &
    \includegraphics[width=0.15\textwidth]{IHIL/ROE2/baseline/056_pred.png} &
    \includegraphics[width=0.15\textwidth]{IHIL/ROE2/baseline/056_ref.png} \\
\vspace{-0.6cm}\centering \small Ours \vspace{0.6cm} &  \includegraphics[width=0.15\textwidth]{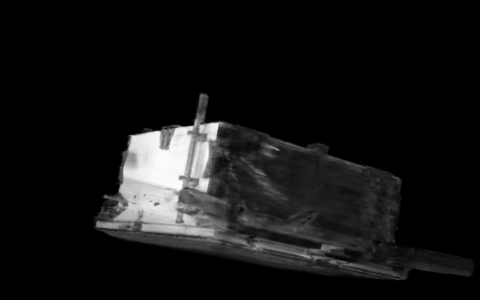} &
    \includegraphics[width=0.15\textwidth]{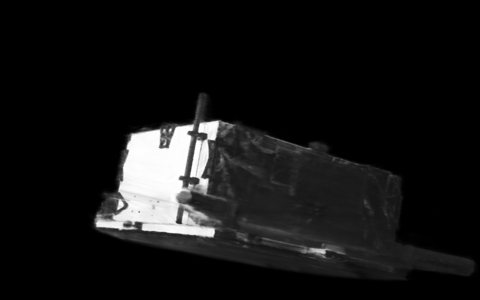} &
    \includegraphics[width=0.15\textwidth]{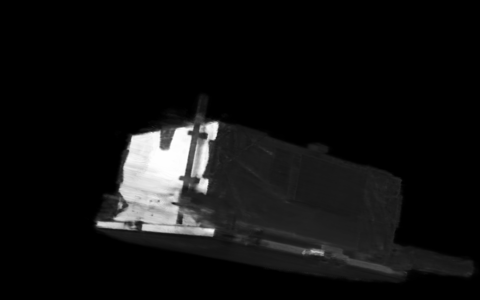} &
    \includegraphics[width=0.15\textwidth]{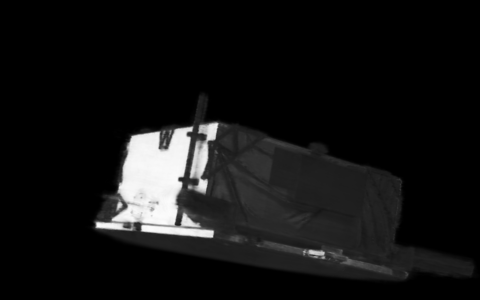} &
    \includegraphics[width=0.15\textwidth]{IHIL/ROE2/itw_wpf/056_pred.png} &
    \includegraphics[width=0.15\textwidth]{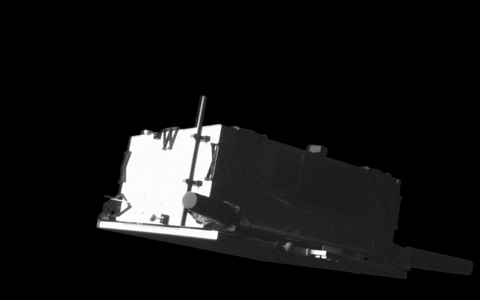} \\
\vspace{-0.6cm}\centering \small Baseline \vspace{0.6cm} & \includegraphics[width=0.15\textwidth]{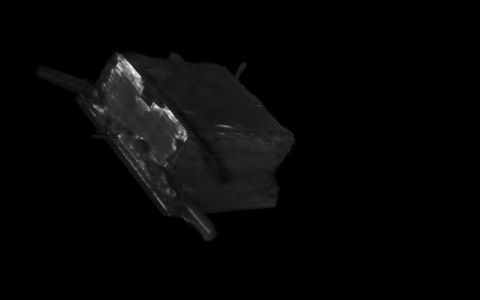} &
    \includegraphics[width=0.15\textwidth]{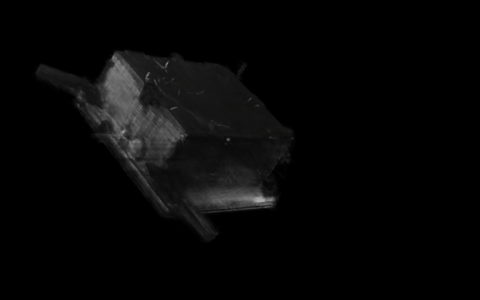} &
    \includegraphics[width=0.15\textwidth]{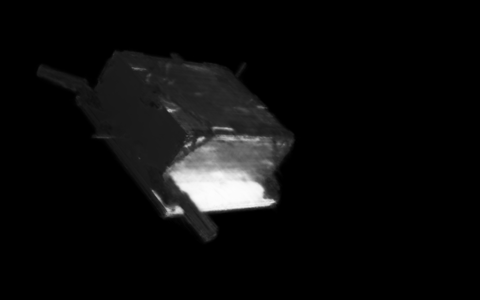} &
    \includegraphics[width=0.15\textwidth]{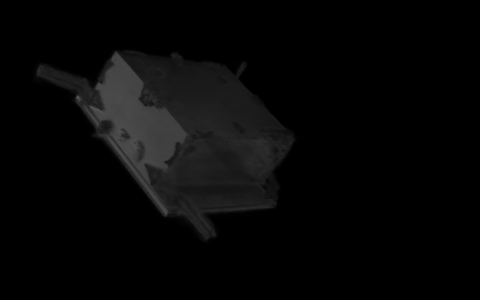} &
    \includegraphics[width=0.15\textwidth]{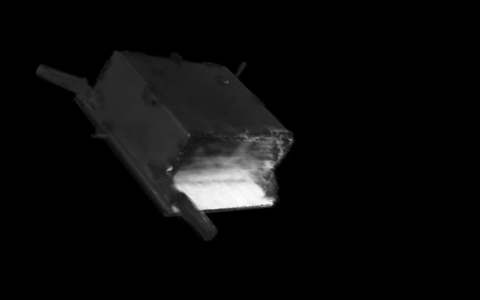} &
    \includegraphics[width=0.15\textwidth]{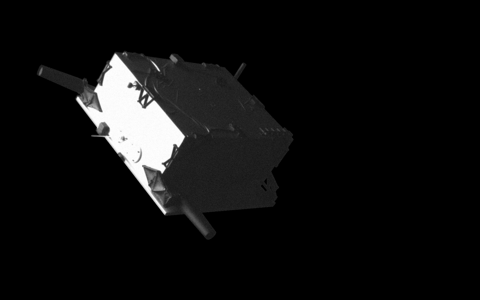} \\
\vspace{-0.6cm}\centering \small Ours \vspace{0.6cm} &  \includegraphics[width=0.15\textwidth]{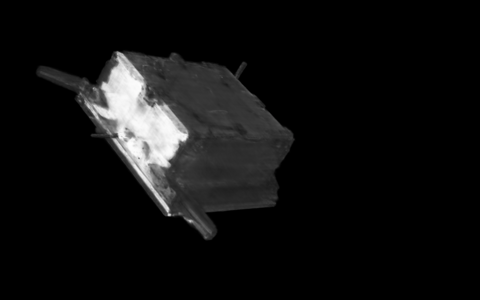} &
    \includegraphics[width=0.15\textwidth]{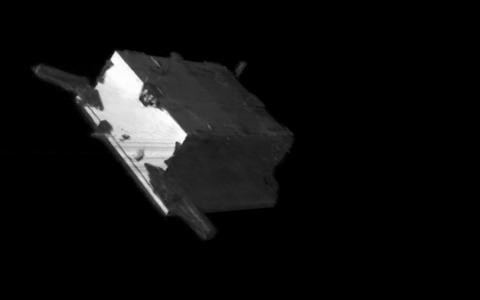} &
    \includegraphics[width=0.15\textwidth]{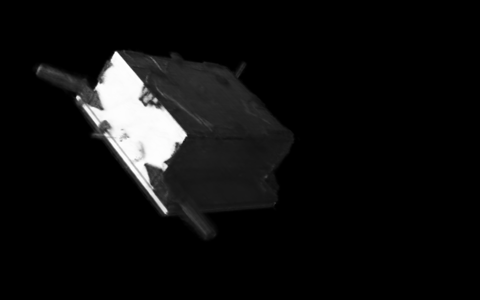} &
    \includegraphics[width=0.15\textwidth]{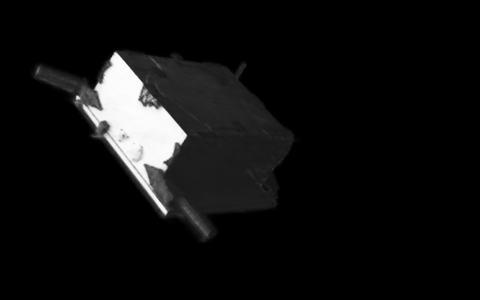} &
    \includegraphics[width=0.15\textwidth]{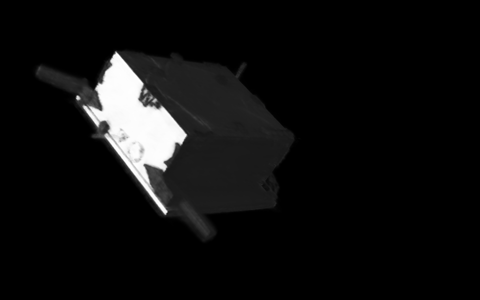} &
    \includegraphics[width=0.15\textwidth]{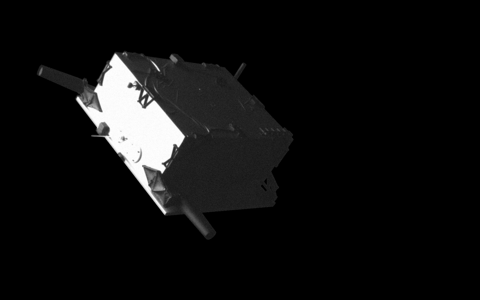} \\
\end{NiceTabular}
\end{table}

\setlength{\tabcolsep}{0pt} 
\renewcommand{\arraystretch}{0} 
\begin{table}[t]
\centering
\caption{\small\label{tab_nimages_sunlamp} Qualitative assessment of our reconstruction method applied on \textit{Sunlamp} images, compared to the baseline reconstruction, for different number of training images. Our method is more resilient to scarce training sets, even on images exhibiting adverse illumination conditions.}
\begin{NiceTabular}{m{1.5cm}@{}*{6}{c@{}}}[hvlines]
\vspace{0.1cm} \, \# images \vspace{0.2cm}  & 25 & 50 & 100 & 200 & 400 & \small Reference\\
\vspace{-0.6cm}\centering \small Baseline \vspace{0.6cm} & \includegraphics[width=0.15\textwidth]{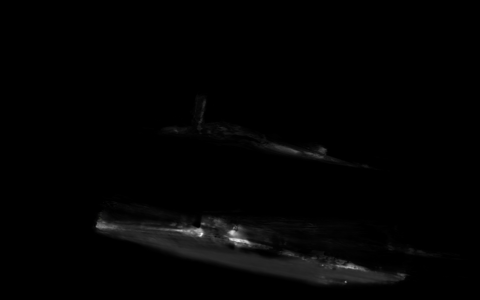} &
    \includegraphics[width=0.15\textwidth]{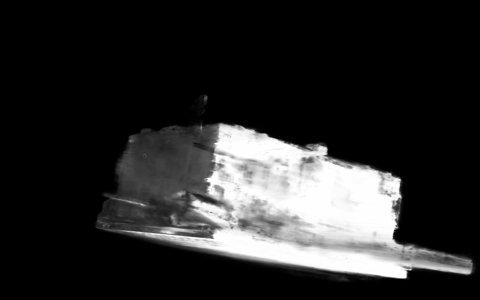} &
    \includegraphics[width=0.15\textwidth]{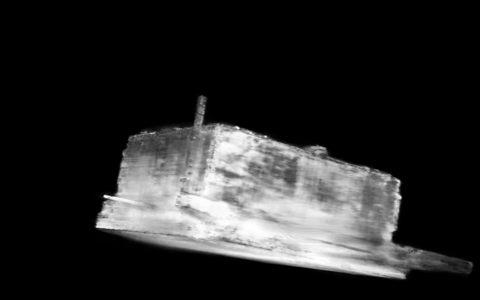} &
    \includegraphics[width=0.15\textwidth]{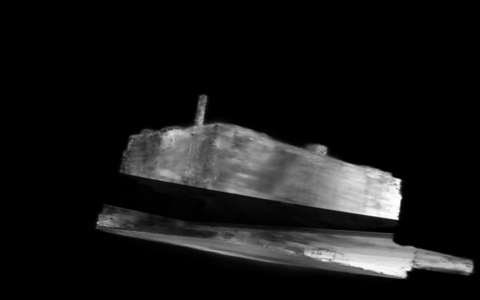} &
    \includegraphics[width=0.15\textwidth]{IHIL/Sunlamp/baseline/056_pred.png} &
    \includegraphics[width=0.15\textwidth]{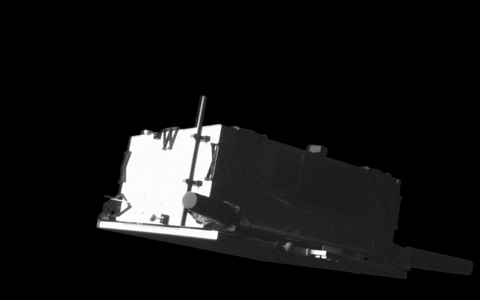} \\
\vspace{-0.6cm}\centering \small Ours \vspace{0.6cm} & \includegraphics[width=0.15\textwidth]{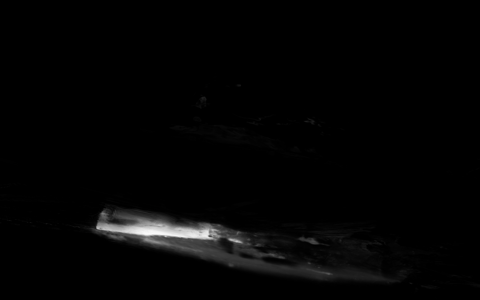} &
    \includegraphics[width=0.15\textwidth]{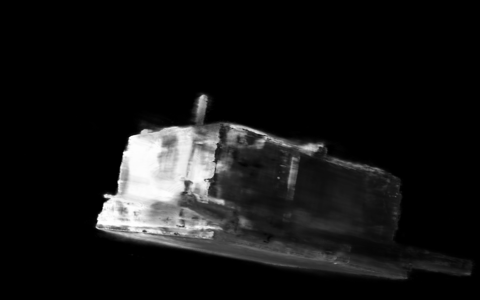} &
    \includegraphics[width=0.15\textwidth]{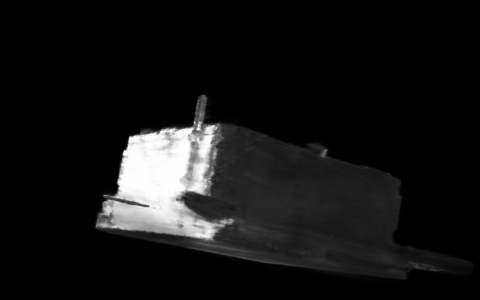} &
    \includegraphics[width=0.15\textwidth]{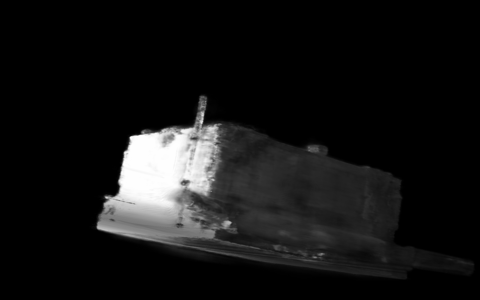} &
    \includegraphics[width=0.15\textwidth]{IHIL/Sunlamp/itw_wpf/056_pred.png} &
    \includegraphics[width=0.15\textwidth]{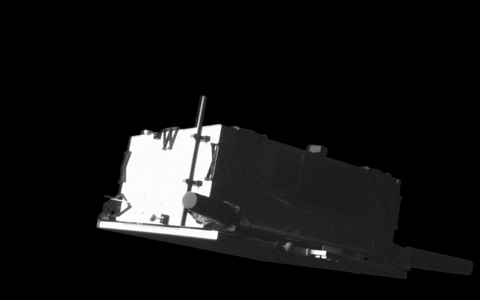} \\
\vspace{-0.6cm}\centering \small Baseline \vspace{0.6cm} & \includegraphics[width=0.15\textwidth]{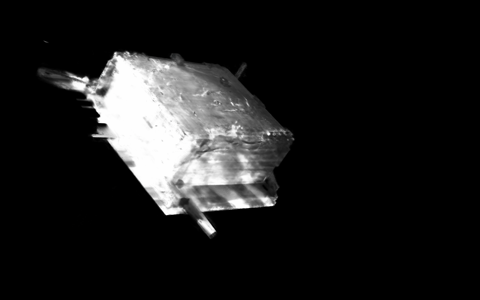} &
    \includegraphics[width=0.15\textwidth]{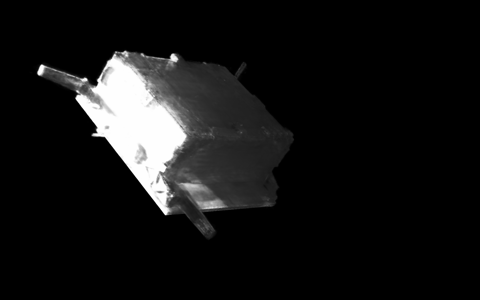} &
    \includegraphics[width=0.15\textwidth]{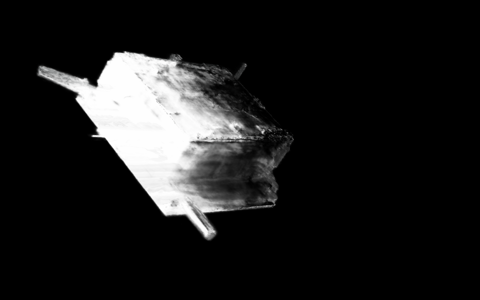} &
    \includegraphics[width=0.15\textwidth]{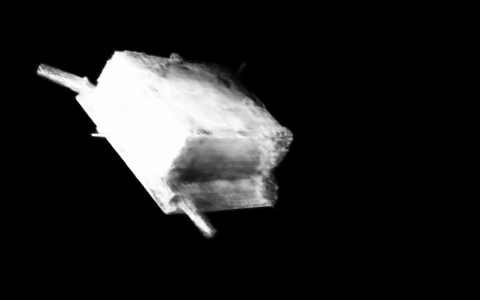} &
    \includegraphics[width=0.15\textwidth]{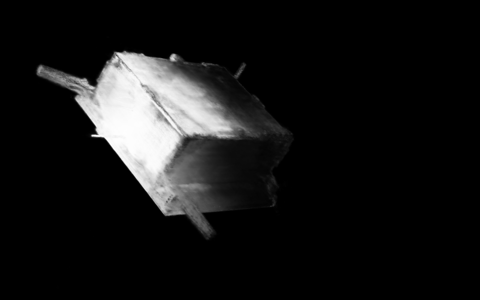} &
    \includegraphics[width=0.15\textwidth]{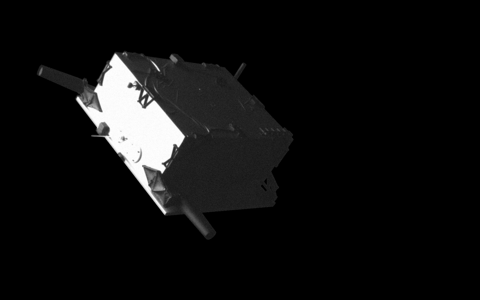} \\
\vspace{-0.6cm}\centering \small Ours \vspace{0.6cm} & \includegraphics[width=0.15\textwidth]{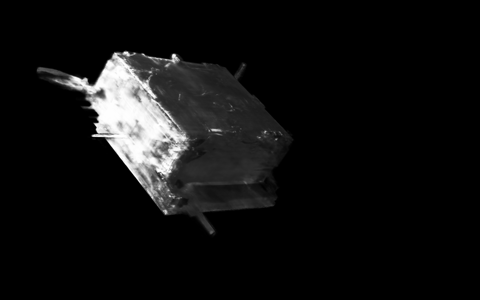} &
    \includegraphics[width=0.15\textwidth]{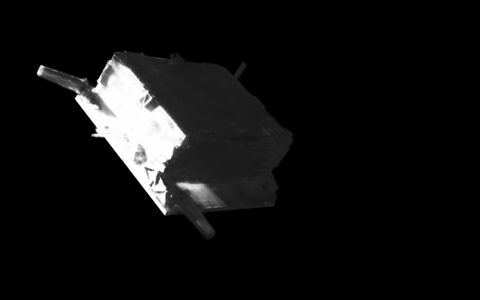} &
    \includegraphics[width=0.15\textwidth]{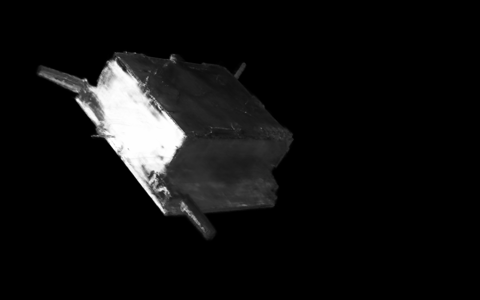} &
    \includegraphics[width=0.15\textwidth]{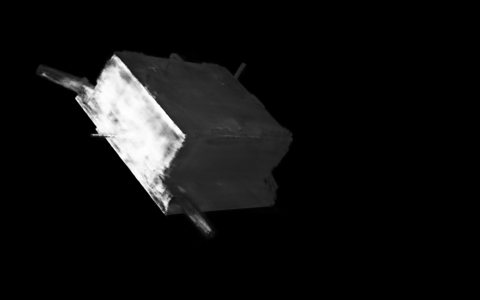} &
    \includegraphics[width=0.15\textwidth]{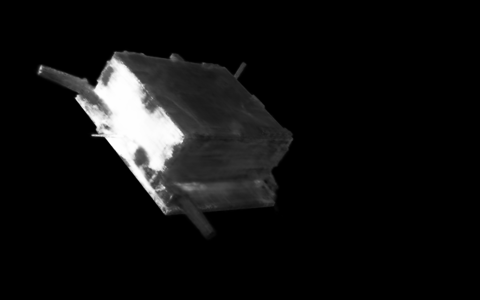} &
    \includegraphics[width=0.15\textwidth]{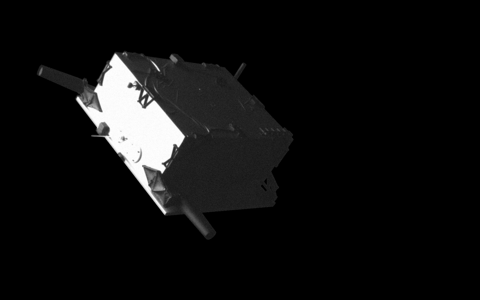} \\
\end{NiceTabular}
\end{table}

To further evaluate robustness under challenging illumination, we repeated experiments on Sunlamp, which simulates direct solar lighting. As shown in \Cref{tab_validation_sunlamp}, the baseline fails to handle these conditions, producing only coarse approximations of the target shape. In contrast, our method reconstructs a consistent representation, recovering overall geometry and subtle details, though less perfectly than on \textit{Lightbox}. Quantitatively, the baseline achieves an average PSNR of 18.84 dB and SSIM of 0.932, while our method reaches an average PSNR of 27.98 dB and SSIM of 0.951.

Our pose correction mechanism on Sunlamp is not as effective as on \textit{Lightbox}, likely because its benefits depend on a sufficiently accurate underlying representation. With pose correction, the average PSNR decreases slightly, from 27.97 dB to 27.14 dB, while the SSIM remains unchanged. However, the JNB score improves from 0.0402 to 0.0389, a 3.2\% reduction, consistent with previous observations on \textit{Lightbox}. Despite minor reductions of the image quality metrics are slightly reduced, pose finetuning enhances image sharpness.

\subsection{Sensitivity to Training Image Count}
\label{sec_experiments_nimages}

\Cref{tab_nimages_roe2_lightbox} illustrates images generated by both the baseline and our method when trained with varying numbers of \textit{Lightbox} images. Both approaches reconstruct geometrically accurate models even with fewer training images. However, while our method preserves the target spacecraft’s appearance under limited data, the baseline struggles as the image count decreases. 
Overall, image quality for our method degrades moderately with fewer training images: the average PSNR, and SSIM drop from 30.54 dB and 0.950 at 400 training images to 27.46 dB and 0.933 at 25 images. 
The JNB score evolves from 0.0337 to 0.0496, which indicates that a reconstructed model is not as sharp when the number of images decreases.
This confirms that performance declines with reduced data, though the degradation remains reasonable.

A similar trend is observed on \textit{Sunlamp}, as illustrated in \Cref{tab_nimages_sunlamp}, where the effect of reduced image count is even more pronounced. The baseline trained on 400 images captures geometry and partially reproduces appearance. However, with fewer images, it only approximates the target shape. In contrast, our method not only performs better with 400 images but also exhibits greater robustness to data scarcity. With 50 images or more, it reconstructs both geometry and appearance. The JNB score increases from 0.0389 at 400 images to 0.0440 at 50 images. 
This is consistent with the observations made on \textit{Lightbox}. 
The reconstruction quality reduces with less images, but the degradation is limited.

\subsection{Ablation Study: Pose Correction}
\label{sec_experiments_finetuning}

\setlength{\tabcolsep}{0pt} 
\renewcommand{\arraystretch}{0} 
\begin{table}[p]
\centering
\caption{\small\label{tab_finetuning_roe2_synthetic_ER} Images generated by our model trained either using pseudo-pose labels or finetuned labels. Each line depicts an image taken under the same pose, for the same model trained on image sets with different level of noise on their relative orientation labels.} 
\begin{NiceTabular}{m{1.7cm}*{5}{c@{}}}[hvlines]
\centering \vspace{0.1cm} \small Pose Finetuning \vspace{0.05cm} &  $E_{\textnormal{R}}$ = 0.0$^\circ$ &  $E_{\textnormal{R}}$ = 0.2$^\circ$ &  $E_{\textnormal{R}}$ = 0.4$^\circ$ & $E_{\textnormal{R}}$ = 0.8$^\circ$ & $E_{\textnormal{R}}$ = 1.6$^\circ$\\
\centering \vspace{-1.2cm} \Large \xmark \vspace{0.7cm} & 
    \includegraphics[width=0.18\textwidth]{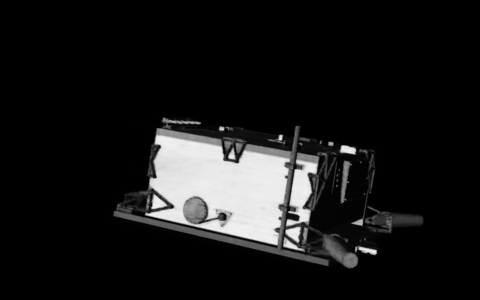} &
    \includegraphics[width=0.18\textwidth]{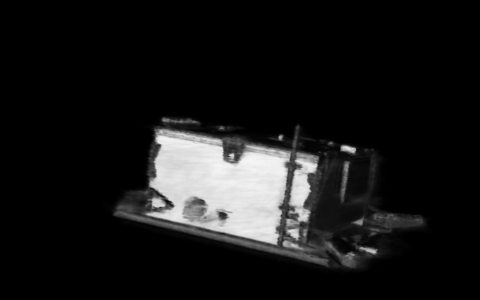} &
    \includegraphics[width=0.18\textwidth]{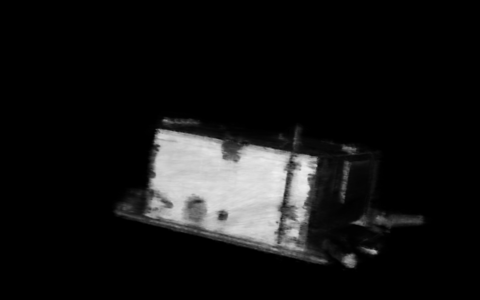} &
    \includegraphics[width=0.18\textwidth]{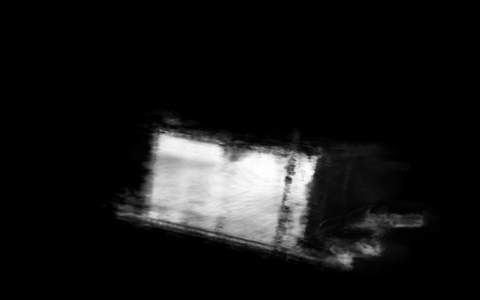} & 
    \includegraphics[width=0.18\textwidth]{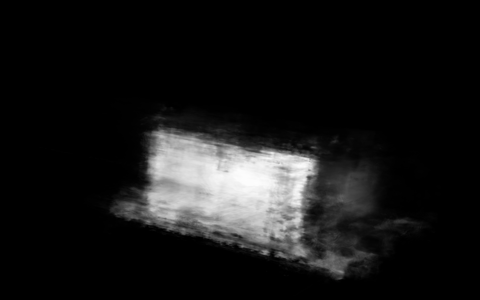} \\
\centering \vspace{-1.2cm} \Large \vmark \vspace{0.7cm} & 
    \includegraphics[width=0.18\textwidth]{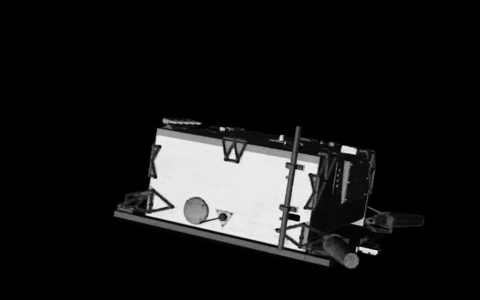} &
    \includegraphics[width=0.18\textwidth]{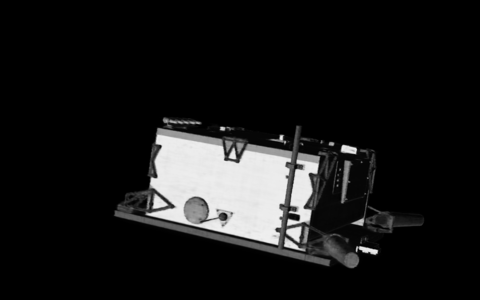} &
    \includegraphics[width=0.18\textwidth]{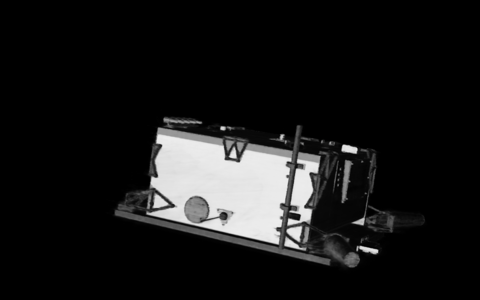} &
    \includegraphics[width=0.18\textwidth]{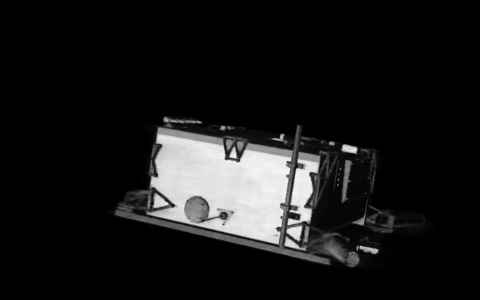} &
    \includegraphics[width=0.18\textwidth]{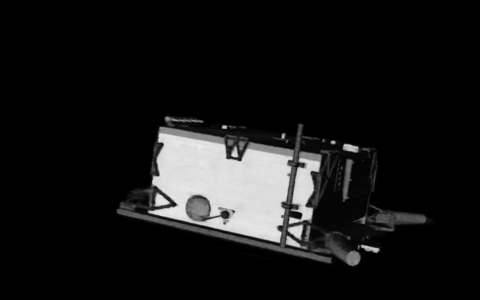} \\
\centering \vspace{-1.2cm} \Large \xmark \vspace{0.7cm} & 
    \includegraphics[width=0.18\textwidth]{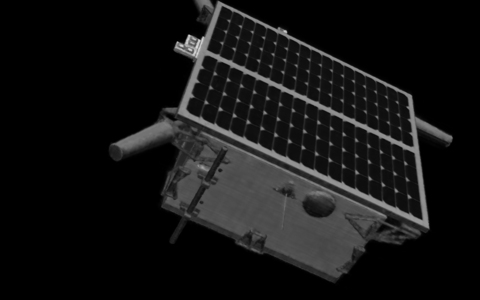} &
    \includegraphics[width=0.18\textwidth]{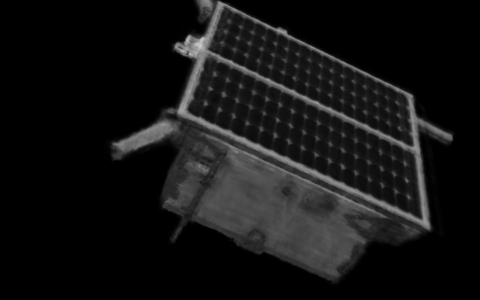} &
    \includegraphics[width=0.18\textwidth]{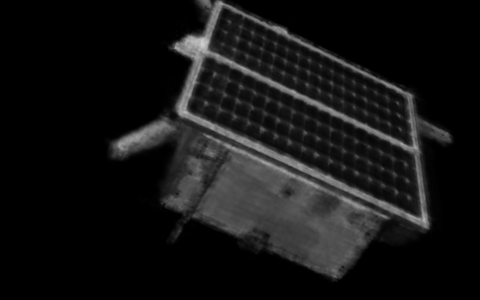} &
    \includegraphics[width=0.18\textwidth]{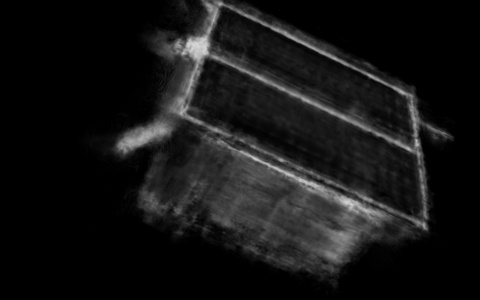} &
    \includegraphics[width=0.18\textwidth]{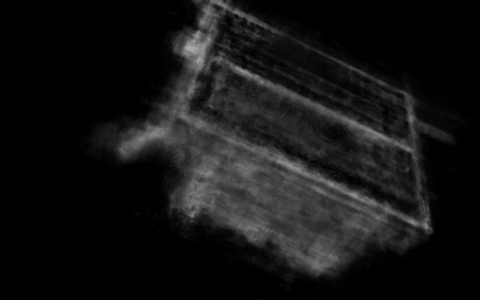} \\
\centering \vspace{-1.2cm} \Large \vmark \vspace{0.7cm} & 
    \includegraphics[width=0.18\textwidth]{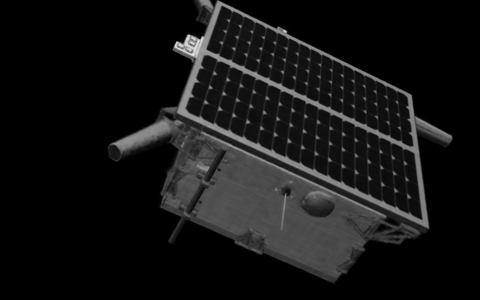} &
    \includegraphics[width=0.18\textwidth]{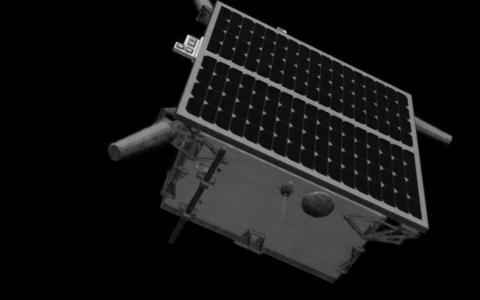} &
    \includegraphics[width=0.18\textwidth]{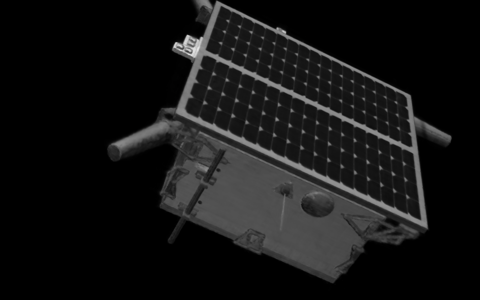} &
    \includegraphics[width=0.18\textwidth]{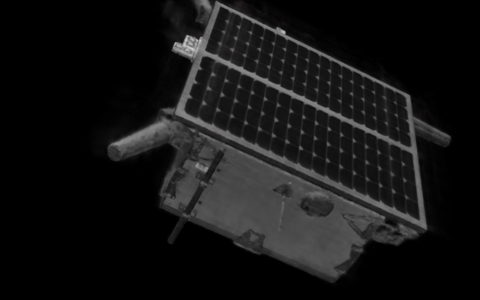} &
    \includegraphics[width=0.18\textwidth]{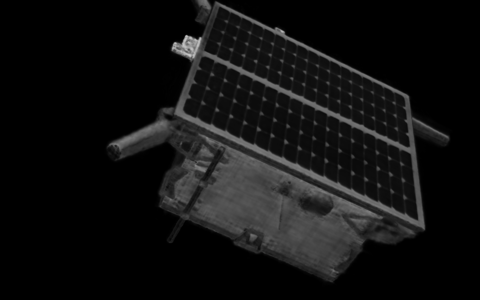} \\
& & &   \\
\end{NiceTabular}
\end{table}

\setlength{\tabcolsep}{0pt} 
\renewcommand{\arraystretch}{0} 
\begin{table}[p]
\centering
\caption{\small\label{tab_finetuning_roe2_synthetic_ET} Images generated by our model trained either using pseudo-pose labels or finetuned labels. Each line depicts an image taken under the same pose, for the same model trained on image sets with different level of noise on their relative position labels.}
\begin{NiceTabular}{m{1.7cm}*{5}{c@{}}}[hvlines]
\centering \vspace{0.1cm} \small Pose Finetuning \vspace{0.05cm} & $E_{\textnormal{T}}$ = 0 mm & $E_{\textnormal{T}}$ = 2 mm & $E_{\textnormal{T}}$ = 4 mm & $E_{\textnormal{T}}$ = 8 mm & $E_{\textnormal{T}}$ = 16 mm \\
\centering \vspace{-1.2cm} \Large \xmark \vspace{0.7cm} & 
    \includegraphics[width=0.18\textwidth]{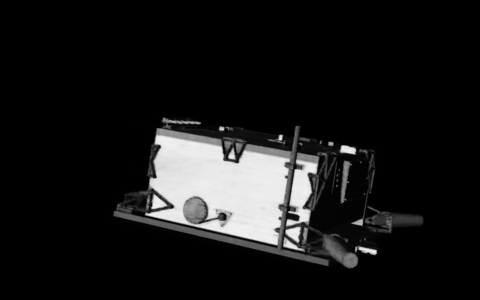} &
    \includegraphics[width=0.18\textwidth]{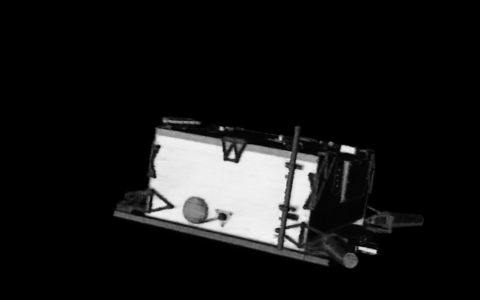} &
    \includegraphics[width=0.18\textwidth]{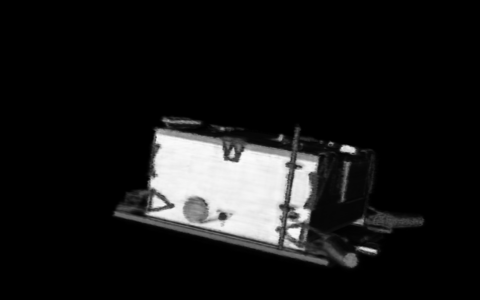} &
    \includegraphics[width=0.18\textwidth]{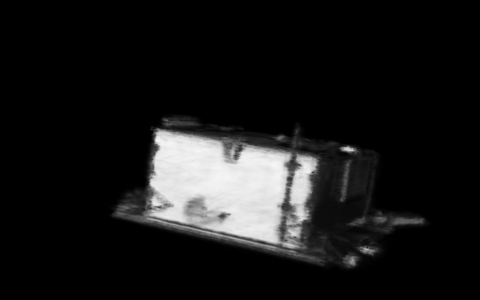} &
    \includegraphics[width=0.18\textwidth]{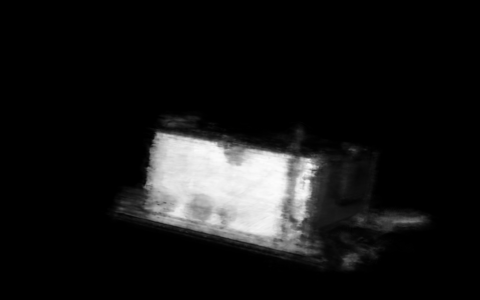} \\
\centering \vspace{-1.2cm} \Large \vmark \vspace{0.7cm} & 
    \includegraphics[width=0.18\textwidth]{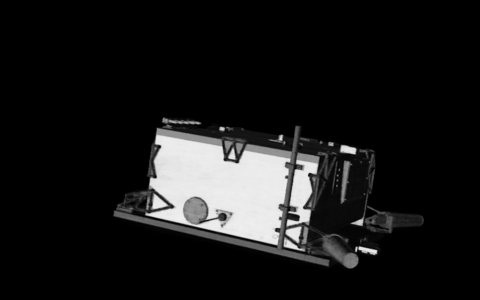} &
    \includegraphics[width=0.18\textwidth]{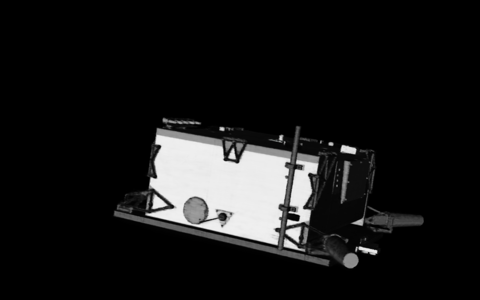} &
    \includegraphics[width=0.18\textwidth]{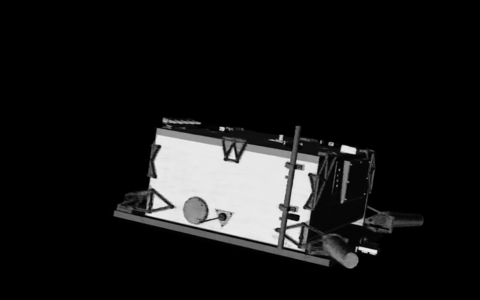} &
    \includegraphics[width=0.18\textwidth]{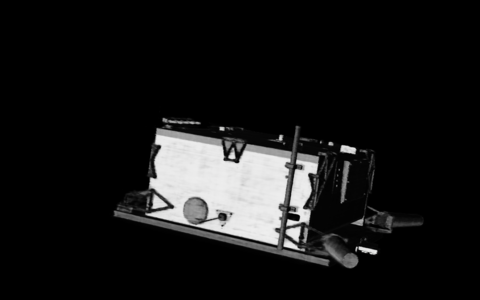} &
    \includegraphics[width=0.18\textwidth]{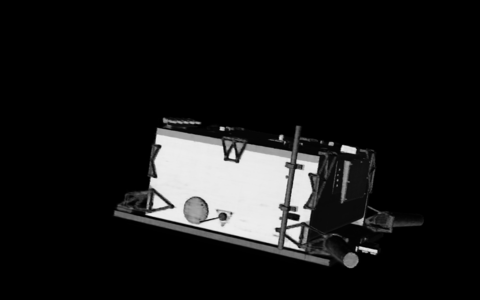} \\
\centering \vspace{-1.2cm} \Large \xmark \vspace{0.7cm} & 
    \includegraphics[width=0.18\textwidth]{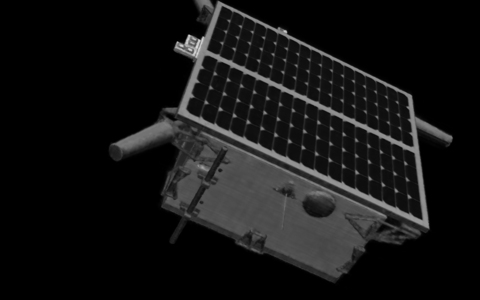} &
    \includegraphics[width=0.18\textwidth]{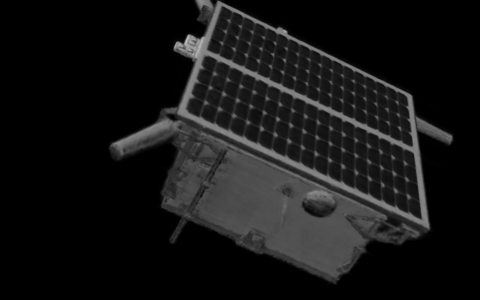} &
    \includegraphics[width=0.18\textwidth]{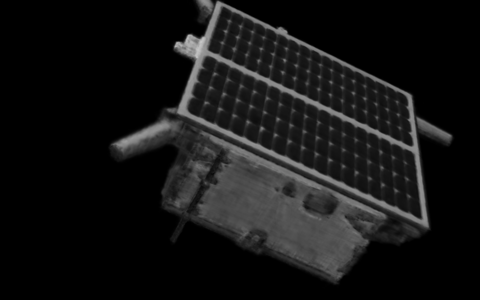} &
    \includegraphics[width=0.18\textwidth]{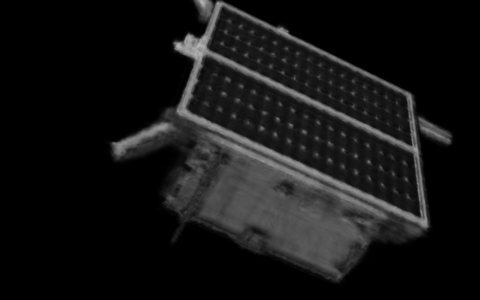} &
    \includegraphics[width=0.18\textwidth]{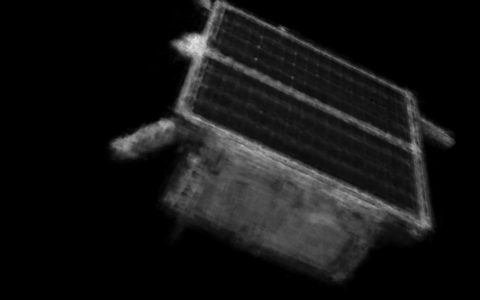} \\
\centering \vspace{-1.2cm} \Large \vmark \vspace{0.7cm} &
    \includegraphics[width=0.18\textwidth]{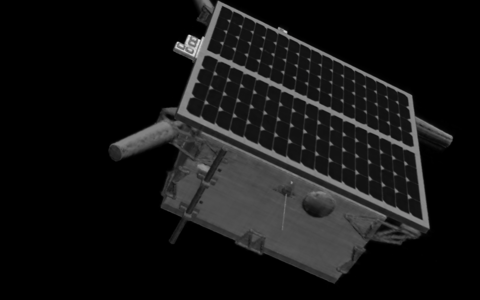} &
    \includegraphics[width=0.18\textwidth]{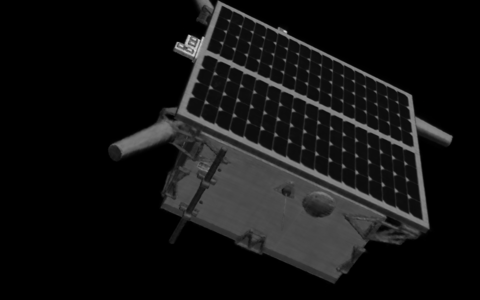} &
    \includegraphics[width=0.18\textwidth]{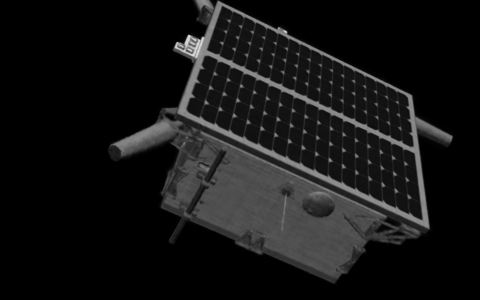} &
    \includegraphics[width=0.18\textwidth]{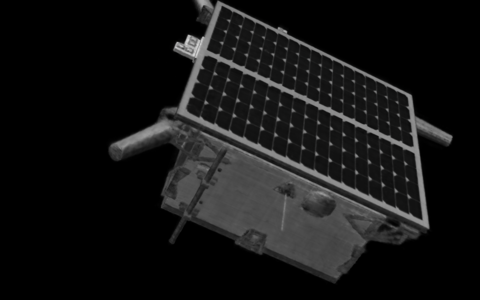} &
    \includegraphics[width=0.18\textwidth]{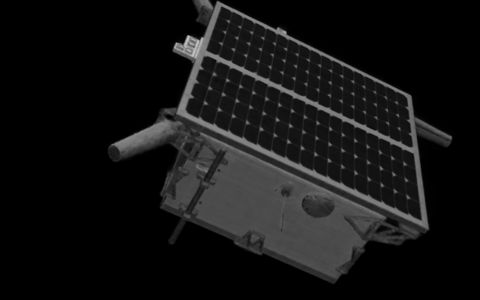} \\
\end{NiceTabular}
\end{table}

\begin{figure}[t]
    \centering
    \includegraphics[width=0.79\textwidth]{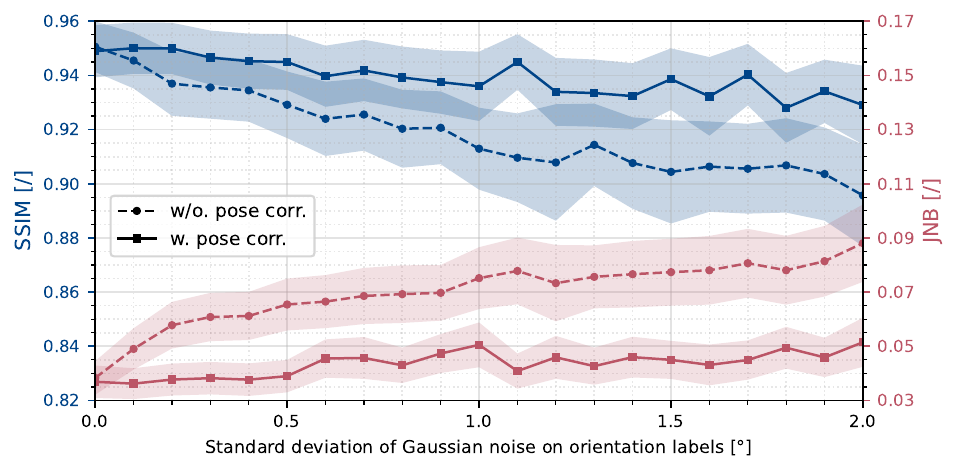}
    \vspace{-0.4cm}
    \caption{\label{fig_pose_finetuning_ER} Relationship between the average angular error on the pose labels and the average image quality metrics (SSIM: higher is better, JNB: lower is better), with or without performing pose correction. Shaded areas correspond to $[\mu - 0.2 \: \sigma, \mu + 0.2  \: \sigma]$. Finetuning the pose labels improves both the average metrics and reduces their deviation.}
    \vspace{-0.4cm}
\end{figure}

\begin{figure}[b!]
    \centering
    \includegraphics[width=0.79\textwidth]{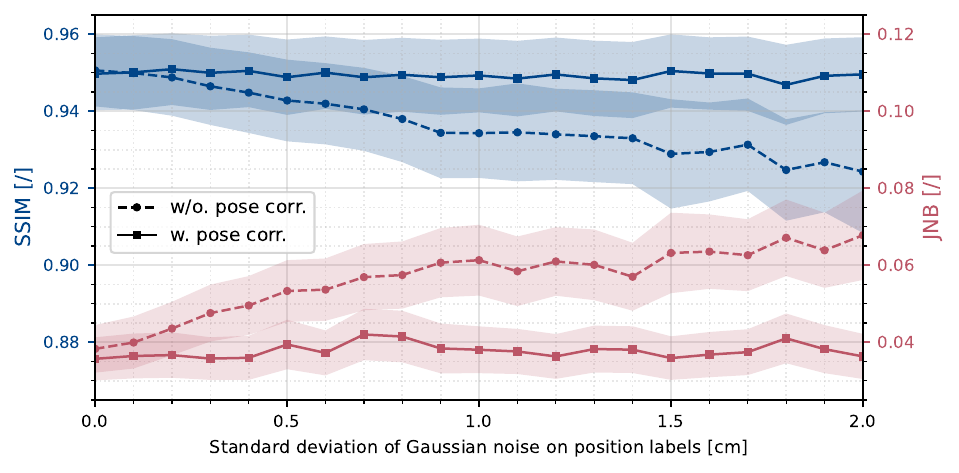}
    \vspace{-0.4cm}
    \caption{\label{fig_pose_finetuning_ET} Relationship between the average position error  on the pose labels and the average image quality metrics, with or without performing pose correction. Shaded areas correspond to $[\mu - 0.2 \: \sigma, \mu + 0.2  \: \sigma]$. Finetuning the pose labels improves both the average metrics and reduces their deviation.}
\end{figure}

To further assess the effectiveness of the proposed pose correction mechanism and its robustness across varying noise levels, we conduct an ablation study. For this purpose, we use the \textit{Synthetic} dataset, which contains noise-free pose labels, enabling a controlled evaluation of noise impact on reconstruction quality. In the following experiments, we introduce synthetic noise to pose labels and train our method with and without pose correction to quantify its influence on reconstruction accuracy.

\Cref{tab_finetuning_roe2_synthetic_ER} shows images generated from models trained on orientation-corrupted labels, with and without pose correction. Reconstructions without pose correction appear blurry, whereas those trained with pose correction are significantly sharper and therefore often match the reference image. For orientation noise up to 0.8$^\circ$, reconstructions are visually perfect; at 1.6$^\circ$, minor artifacts appear. These results indicate that pose correction effectively compensates for substantial orientation noise, well beyond typical annotation uncertainty. \Cref{fig_pose_finetuning_ER} illustrates the relationship between noise standard deviation and image quality metrics for models trained with and without pose correction. Across all noise levels, pose correction consistently improves reconstruction quality in terms of both image quality, \ie, SSIM, and sharpness, \ie,JNB score.

Similarly, \Cref{tab_finetuning_roe2_synthetic_ET} presents the results for models trained on position-corrupted labels. As with orientation noise, pose correction significantly mitigates the impact of position label uncertainty, producing sharper representations. At low noise levels, reconstructions from pose-corrected models are virtually artifact-free; at higher levels, only minor artifacts remain. \Cref{fig_pose_finetuning_ET} confirms these observations quantitatively: pose correction improves reconstruction quality across all position noise levels, reinforcing its effectiveness in handling pose uncertainty.

\section{Conclusion}
\label{sec_conclusion}

    This paper proposes a method for reconstructing a 3D model of an unknown spacecraft from monocular 2D imagery. The approach specifically addresses two critical challenges often overlooked by state-of-the-art techniques when applied to datasets representative of operational scenarios. These challenges consist of significant illumination variability that undermines the quality and consistency of on-orbit imagery and the pose uncertainty that reduces the reconstruction accuracy.

    To overcome these challenges, our method extends Neural Radiance Fields with additional per-image degrees of freedom. For each image, it learns alongside the NeRF an appearance embedding that implicitly captures image-specific illumination conditions and a pose correction term that refines the corresponding noisy pose label. We demonstrated that despite their minimal impact on complexity, these additional parameters substantially enhance robustness to both illumination variability and pose uncertainty on a dataset representative of orbital conditions.
    
    Beyond nominal conditions, we also evaluated the method’s robustness in three challenging scenarios: limited image availability (\Cref{sec_experiments_nimages}), severe illumination variations (\Cref{tab_validation_sunlamp}), and increased pose uncertainty (\Cref{sec_experiments_finetuning}). These experiments demonstrate that our approach outperforms prior methods under all these conditions, thereby confirming the effectiveness of its two key components, namely, the appearance embeddings and the learnable pose correction terms.

\section*{Funding Sources}
The research was funded by Aerospacelab and the Walloon Region through the Win4Doc program. C. De Vleeschouwer is a Research Director of the Fonds de la Recherche Scientifique – FNRS.

\section*{Acknowledgments}
Computational resources have been provided by the supercomputing facilities of the Université catholique de Louvain (CISM/UCL) and the Consortium des Équipements de Calcul Intensif en Fédération Wallonie Bruxelles (CÉCI) funded by the Fond de la Recherche Scientifique de Belgique (F.R.S.-FNRS) under convention 2.5020.11 and by the Walloon Region. 

The authors used MS Copilot during the preparation of this manuscript to assist with figure generation from textual descriptions, suggest alternative phrasings, and perform minor language corrections. All scientific content, methodology, and conclusions were developed and validated by the authors, who take full responsibility for the work.

\bibliography{bib}

\end{document}